\documentclass[lettersize,journal]{IEEEtran}
\usepackage{amsmath,amsfonts}
\usepackage{amssymb}
\usepackage{amsmath}
\usepackage{algorithmic}
\usepackage{url}
\usepackage{tabularx}
\usepackage{booktabs}
\usepackage{subfigure}
\usepackage{graphicx}
\usepackage{cite}
\newtheorem{definition}{Definition}

\usepackage{url}
\usepackage{doi}
\usepackage[ruled,vlined]{algorithm2e} 
\usepackage{multirow}
\usepackage{color}
\newenvironment{sloppypar*}
{\sloppy\ignorespaces}
{\par}

\begin{document}\sloppy

\title{Poisoning Semi-supervised Federated Learning via Unlabeled Data: Attacks and Defenses}

\author{Yi Liu, Xingliang Yuan, Ruihui Zhao, Cong Wang,~\IEEEmembership{Fellow,~IEEE}, Dusit Niyato,~\IEEEmembership{Fellow,~IEEE}, and Yefeng Zheng,~\IEEEmembership{Fellow,~IEEE}
\thanks{Yi Liu and Cong Wang are with the Department of Computer Science, City University of Hong Kong, Hong Kong, China (e-mail: yiliu247@cityu.edu.hk; congwang@cityu.edu.hk). Xingliang Yuan is with the Faculty of Information Technology, Monash University, Clayton, VIC 3800, Australia (e-mail: xingliang.yuan@monash.edu). Ruihui Zhao and Yefeng Zheng are with the Tencent Jarvis Lab, Tencent, China (e-mail: zacharyzhao@tencent.com; yefengzheng@tencent.com). Dusit Niyato is with School of Computer Science and Engineering (SCSE), Nanyang Technological University, Singapore (e-mail: dniyato@ntu.edu.sg).}}

\markboth{IEEE Transactions on Information Forensics and Security}%
{Shell \MakeLowercase{\textit{et al.}}: A Sample Article Using IEEEtran.cls for IEEE Journals}


\maketitle

\begin{abstract}
Semi-supervised Federated Learning (SSFL) has recently drawn much attention due to its practical consideration, i.e., the clients may only have unlabeled data. In practice, these SSFL systems implement semi-supervised training by assigning a ``guessed'' label to the unlabeled data near the labeled data to convert the unsupervised problem into a fully supervised problem. However, the inherent properties of such semi-supervised training techniques create a new attack surface.
In this paper, we discover and reveal a simple yet powerful poisoning attack against SSFL. Our attack utilizes the natural characteristic of semi-supervised learning to cause the model to be poisoned by poisoning unlabeled data. Specifically, the adversary just needs to insert a small number of maliciously crafted unlabeled samples (e.g., only 0.1\% of the dataset) to infect model performance and misclassification. Extensive case studies have shown that our attacks are effective on different datasets and common semi-supervised learning methods. To mitigate the attacks, we propose a defense, i.e., a minimax optimization-based client selection strategy, to enable the server to select the clients who hold the correct label information and high-quality updates. Our defense further employs a quality-based aggregation rule to strengthen the contributions of the selected updates. Evaluations under different attack conditions show that the proposed defense can well alleviate such unlabeled poisoning attacks. Our study unveils the vulnerability of SSFL to unlabeled poisoning attacks and provides the community with potential defense methods.

\end{abstract}

\maketitle

\section{Introduction}
Federated Learning (FL) \cite{ref-5} is an emerging decentralized training paradigm that allows clients (e.g., mobile phones) to jointly train a shared global model \cite{ref-8,ref-31,weng2021fedserving,zhang2020enabling,zheng2022aggregation}. In FL, clients are asked to upload the model updates to a server while keeping their data locally during training, which prevents them from sharing the raw data for privacy protection \cite{ref-6,zhu2021semi,liu2022right}. 
Note that most FL-based systems are based on a restrictive assumption; that is, the local datasets of the clients are labeled. In the real world, however, it is not common for each client to have rich labeled data. For example, during the COVID-19 epidemic, inter-geographic hospitals might not be able to train a high-accuracy diagnostic model because some of them do not have enough labeled data. Furthermore, putting together a properly labeled dataset could be a time-consuming, expensive, and complicated endeavor \cite{ref-2}. In the above example, annotating lung X-rays requires professional doctors with specific domain knowledge and consumes much time.

To overcome the above limitations, recent work \cite{ref-37,ref-38,wang2020graphfl,lin2021semifed} made exploration on the design of the semi-supervised federated learning (SSFL) system, which focuses on how to apply the semi-supervised techniques to FL. Indeed, the integration of semi-supervised learning and FL addresses this by allowing a global model to be trained on a small set of (\textit{expensive-to-collect}) labeled samples and a large set of (\textit{cheap-to-collect}) unlabeled samples \cite{carlini2021poisoning}. 

While SSFL provides users with many benefits, it opens up a new attack surface for adversary. In this work, we are interested in poisoning attacks on \textit{unlabeled data} in the SSFL setting. Such critical attacks cause the model to fool itself into mislabeling arbitrary input samples. As known, existing semi-supervised training techniques rely heavily on the pseudo-label information of unlabeled data \cite{kingma2014semi,zhai2019s4l}. Specifically, semi-supervised federated learning allows the server to hold a small amount of labeled data and train a model on it that can assign pseudo-labels to the unlabeled data on the client. In this way, the unsupervised training of the client on unlabeled data is transformed into supervised training on pseudo-labeled data. In light of the above fact, we observe that the unlabeled data of semi-supervised learning provide adversaries with an attack surface. An adversary can inject carefully-selected samples or manipulate unlabeled data, causing the learned model to misclassify input samples or lose accuracy. 




Different from poisoning attacks against the supervised FL (SFL)~\cite{9524709,9285303}, we discover that an adversary can exclusively poison the unlabeled data. Standing on top of it, we emphasize that our poisoning attacks are different from those in SFL, due to the following scenario specific reasons.
%
%
In semi-supervised learning, the classifier normally unconditionally trusts the pseudo-labels assigned to the unlabeled data. As a result, once the adversary poisons the unlabeled data, the victim classifier will have a classification error or an accuracy loss due to its limitation  to identify the correctness of pseudo-labels. In contrast, in supervised learning, the classifier can utilize some metrics (e.g., test accuracy) to possibly identify poisoned samples.



On the other hand, compared with SFL, SSFL not only completes the original training procedure, but also  uses the classifier on the server to set accurate pseudo-labels for local unlabeled data.
Once the unlabeled data on the client is poisoned, the ability of the classifier on the server to correctly set pseudo-labels will be affected gradually.
Since the true labels of the local unlabeled data are agnostic, the server cannot tell whether the pseudo labels are correct or not. 
This is critical as the correctness of the label information of unlabeled data is closely related to the accuracy of the ``decision boundary'' on which the SSL model classification accuracy depends \cite{berthelot2019mixmatch}.


To address the poisoning attacks in SFL, a few defenses have been proposed~\cite{9551983,9524709,zhang2020enabling,awan2021contra,ref-80,ref-76}. However, most of them can hardly be applicable to SSFL\cite{chen2021cartl} due to the difference between the training procedures of SSFL and SFL. In particular, some efforts have been made to remove poisoned data with the help of label information \cite{9551983}.
But the unlabeled data in SSFL is processed by techniques such as data augmentation \cite{zhong2020random}, label regularization \cite{yuan2020revisiting}, and label propagation \cite{wang2007label}, and in this case, the defense cannot simply rely on the label information to defend against such poisoning attacks. 
%
%
Second, recent advances in SFL assume that unlabeled training data is sampled from the same trusted sources as labeled training data. However, in SSFL, unlabeled training data may come from untrusted sources \cite{9551983}. 
It becomes much more challenging to prevent from poisoning unlabeled data.


In this paper, we present and analyze a simple yet critically harmful poisoning attack against SSFL. Unlike existing poisoning attacks against SFL, the attack focuses on poisoning unlabeled data by utilizing the natural characteristic of semi-supervised learning to achieve high attack performance. Firstly, we design and implement the poisoning attacks in SSFL, and comprehensively evaluate and demonstrate the corresponding vulnerability of SSFL. Secondly, we further use different semi-supervised methods to evaluate the effectiveness of the designed poisoning attacks on various datasets for SSFL. Experimental results show that by manipulating only 0.1\% of unlabeled examples, we can classify mislead the model to a specific target sample into any desired category. Thirdly, we propose a client selection strategy based on the minimax optimization strategy to defend against the proposed poisoning attacks. The intuition behind such defense is to help the model obtain high quality label information and model updates.

The contributions of this paper are summarized as follows:
\begin{itemize}
	\item 
	
	We first design a simple yet powerful poisoning attack in semi-supervised federated learning which only needs to manipulate a small amount (e.g., $0.1\%$ in our experiment) of unlabeled data to achieve high attack performance. Specifically, we discover the inherent properties of semi-supervised learning to make the model fool itself to achieve misclassification.
	
	
	\item 
	We comprehensively evaluate the robustness of a semi-supervised federated learning to deal with the poisoning attacks and reveal its vulnerability to such attacks. In SSFL, the experimental results show that the harm of poisoning unlabeled data is much greater than that in other existing poisoning attacks.
	
	\item 
	We design a minimax optimization-based client selection strategy to defend the attacks. To obtain high-confidence label information and high-quality model updates, our defense minimizes the difference between the client's updates and the server's global model updates while maximizing the cosine similarity and Wasserstein distance between the client's updates and the server's global model updates. Furthermore, we propose a quality-based aggregation rule to aggregate the selected clients' updates robustly. 
	
	\item 
	%
    We conduct an extensive evaluation over two real-world datasets. In particular, we simulate distributed and centralized poisoning scenarios in the first place, and evaluate our attacks and defenses in these scenarios. Comprehensive case studies show that the proposed poisoning attack is effective. Furthermore, we verify the effectiveness of our defenses under different attack settings.
\end{itemize}

\begin{table}
\scriptsize
  \caption{Summary of Main Symbols.}
  \label{table:symbols}
  \centering
  \renewcommand{\arraystretch}{1}
  \begin{tabular}{p{40pt}<{\centering}|p{185pt}<{}}
    \hline
    \textbf{Symbol} & \textbf{Explanation} \\
    \hline
    $c$ & The number of compromised clients \\
    \hline
    $\mathcal{D}_s$ & The small labeled dataset at the server\\
    \hline
    $\mathcal{D}_k$ & The unlabled dataset at the client $k$ \\
    \hline
    $\mathcal{D}_k^p$& The poisoned unlabled dataset of the client $k$\\
    \hline
    $K$ & The total number of clients in SSFL \\
    \hline
    $k$& The identity of client\\
    \hline
    $N_s$ & The number of samples in the labeled dataset at the server $\mathcal{S}$\\
    \hline
    $N_k$ & The number of samples in local unlabeled dataset at the client $k$\\
    \hline
    $p$& The poisoning rate\\
    \hline
    $q$ & The proportion of client participation\\
    \hline
    $\mathcal{S}$ & The server\\
    \hline
    $x_i$& The input samples\\
    \hline
    $y,\hat y,{y^*}$ & The true label, the pseudo label, and the target label\\
    \hline
    $\alpha$ & The threshold of the proposed client selection strategy\\
    \hline
    $\beta$& The threshold to help the server to decide which updates have high confidence to be aggregated\\
    \hline
    $\lambda$ & The threshold hyperparameter of the SSFL\\
    \hline
    ${\lambda _{{\ell_2}}},{\lambda _{{\ell_1}}}$ & The relative weights of $\ell_2$ and $\ell_1$ regularization\\
    \hline
    $\sigma ^2$ & The variance of Gaussian noise\\
    \hline
    $\epsilon$ & The small perturbation\\
    \hline
    $\omega_t$ & The global model's parameters at training round $t$\\
    \hline
    $\omega_t^k$ & The $k$-th client's local model parameters\\
    \hline
    ${\Delta _{i,t}}$ &  The measure of the update quality of the $t$-th round of training \\
    \hline
    $\phi ( \cdot )$ &  The weak data augmentation function\\
    \hline
    $\varphi ( \cdot )$& The strong data augmentation function \\
    \hline
    $\gamma(\cdot)$& The indicator function\\
    \hline
    $Q( \cdot )$ & The quality-based aggregation rule\\
    \hline
    $\ell ( \cdot , \cdot )$& The cross-entropy loss of the client\\
    \hline
    $P( \cdot , \cdot , \cdot )$& The used poisoning function\\
    \hline
  \end{tabular}
  \vspace{-0.5cm}
\end{table}


We highlight that the community can benefit from the designed attacks and defenses: \textit{(i)} This study demonstrates that the attack is practical and can damage broader SSFL-based applications. \textit{(ii)} This study discovers a new attack surface against the learning principles of SSFL, i.e., poisoning unlabeled data. \textit{(iii)} This study provides researchers with a potential defense design that leverages client selection to mitigate such attacks.

The remainder of this paper is organized as follows. We first summarize the related work about poisoning attacks in Sec. \ref{sec-2}. We then review the background of semi-supervised federated learning and introduce our threat model in Sec. \ref{sec-7}. Then, we present the proposed poisoning attacks in Sec. \ref{sec-3}. Next, we elaborate on the proposed defense in Sec. \ref{sec-4}. Subsequently, we conduct a series of case studies on two real-world datasets and analyze the simulation results in Sec. \ref{sec-5}. Finally, we conclude this paper in Sec. \ref{sec-6}. We summarize the mathematical symbols and explanations in Table \ref{table:symbols}.

\section{Related Work}\label{sec-2}
\subsection{Poisoning Attacks}

Poisoning attacks \cite{jagielski2018manipulating} attempt to generate maliciously-crafted data-label pairs to harm the model's classification performance or construct a non-linear mapping path between the target label and the specially designed trigger pattern in the infected model. Various poisoning attacks have been proposed, and they can be divided into untargeted poisoning attacks and targeted poisoning attacks.

\noindent \textbf{Untargeted Poisoning Attacks.} In untargeted poisoning attacks \cite{tolpegin2020data,ref-78,awan2021contra}, the adversary's goal is to degrade the performance of the model. In FL, a straightforward method \cite{ref-78} is to directly manipulate the clients to flip the label of local data to implement this attack. However, previous work efforts have successfully mitigated such attacks by detecting poisoned model updates and malicious clients \cite{awan2021contra}.

\noindent \textbf{Targeted Poisoning Attacks.} Unlike untargeted poisoning attacks, the goals of targeted poisoning attacks \cite{ref-13,ref-75,shafahi2018poison} is to cause the model to output specific (incorrect) predictions for specific samples.
How to design and implement targeted poisoning attacks in SFL has become a research hotspot. For example, Bhagoji \textit{et al.} in \cite{ref-13} designed an adversarial dataset to make the model output specific incorrect predictions for specified samples. In addition, backdoor attacks are sometimes classified as targeted poisoning attacks \cite{ref-15}. Such an attack generally needs to set a trigger on the sample to infect the model. For instance, Bagdasaryan \textit{et al.} in \cite{ref-15} implemented a backdoor attack in the FL setting for the first time. Note that we do not consider backdoor attacks in this paper.

\noindent \textbf{Poisoning Supervised Federated Learning.} Previous work \cite{ref-13,ref-14,ref-15,ref-16} has conducted  research on how to poison supervised federated learning. Generally speaking, they allow a certain number (generally less than $\frac{2}{3}$ of the total) of malicious clients to collude and manipulate and use data-label pairs to poison the dataset to achieve global model performance degradation. Furthermore, they often need to manipulate $1\%  \sim 5\% $  \cite{9551983} of the dataset to achieve decent attack performance. In this paper, our attack confirms that it is not necessary to poison such a large proportion of the data.

\noindent \textbf{Defense Against Poisoning Attacks in FL:} 
To cope with poisoning attacks in FL, several defense methods against poisoning attacks have been developed by mainly using \textit{anomaly-detection-based methods} and \textit{Byzantine-robust aggregation methods} \cite{ref-72,9551983,9524709,zhang2020enabling,awan2021contra,ref-80,ref-76,ref-75,ref-13}. 
%
The anomaly-detection-based methods can be divided into two categories: dimensionality reduction based and neural network based methods. 
In \cite{ref-72}, Wu \textit{et al.} transformed the task of detecting abnormal clients into a high-dimensional binary classification problem.  Shen \textit{et al.}~\cite{ref-75} and Li \textit{et al.}~\cite{ref-76} used the autoencoder and its variants to classify high-dimensional weight vectors for malicious client detection. 
However, the above methods are less practical in real life, because they require the server to implement complex dimensionality reduction techniques or train a large-scale deep learning model. 

Byzantine-robust FL aggregation \cite{ref-13,ref-76,ref-80} is a commonly used method to mitigate poisoning attacks. 
%
The goal therein is to learn an accurate global model when a bounded number of clients are malicious. Their key idea is to leverage Byzantine-robust aggregation rules, which essentially compare the clients' local model updates and remove statistical outliers before using them to update the global model. 
Specifically, Li \textit{et al.} in \cite{ref-76} proposed an aggregation rule called RSA, which uses the geometric median of the client's local update as the global model update. Unfortunately, recent studies \cite{ref-13,ref-14} demonstrated that Byzantine-robust aggregation is still vulnerable to targeted model poisoning attacks.

\noindent \textbf{Poisoning Semi-supervised Learning.} Recently, some researchers have turned their attention to poisoning semi-supervised learning. For example, Carlini \textit{et al.} in~\cite{carlini2021poisoning} proposed an interpolation consistency poisoning attacks against semi-supervised learning. Specifically, they poison a semi-supervised learning model by inserting poisoned unlabeled data that can modify the model's ``decision boundary''. However, the prior art has not yet explored poisoning attacks in SSFL. In this work, we propose a new defense method inspired by~\cite{carlini2021poisoning}.

\subsection{Semi-supervised Federated Learning}

To overcome the limitation of the scarcity of labeled data, how to design a semi-supervised federated learning system draws much attention. The current  methods \cite{ref-11,ref-35,ref-37,ref-38} focus on the  efficient integration between semi-supervised learning techniques (e.g., FixMatch \cite{sohn2020fixmatch}, UDA \cite{xie2020unsupervised}, and MixMatch \cite{berthelot2019mixmatch}) and FL. Specifically, they introduced semi-supervised techniques into FL and designed some efficient semi-supervised training methods such as federated consistency loss \cite{ref-35} and regularization techniques \cite{ref-60} to improve the performance of the model. Indeed, the key idea of these methods is to use semi-supervised techniques to assign pseudo-labels to unlabeled data to construct a ``supervised'' task to train the model. Although the implementation details of those methods are different, they all rely heavily on the ``supervision'' loss that depends on the accuracy of the label information of the unlabeled data. 

On the other hand, there is also a part of work focusing on the extension of the SSFL framework, such as the graph-based SSFL \cite{wang2020graphfl} and the ensemble learning-based SSFL \cite{bian2021fedseal}. However, no one has discussed the security of the SSFL systems, especially in dealing with poisoning attacks caused by poisoning unlabeled data.

\section{Background}\label{sec-7}

\begin{figure}[!t]
	\centering
	\includegraphics[width=1\linewidth]{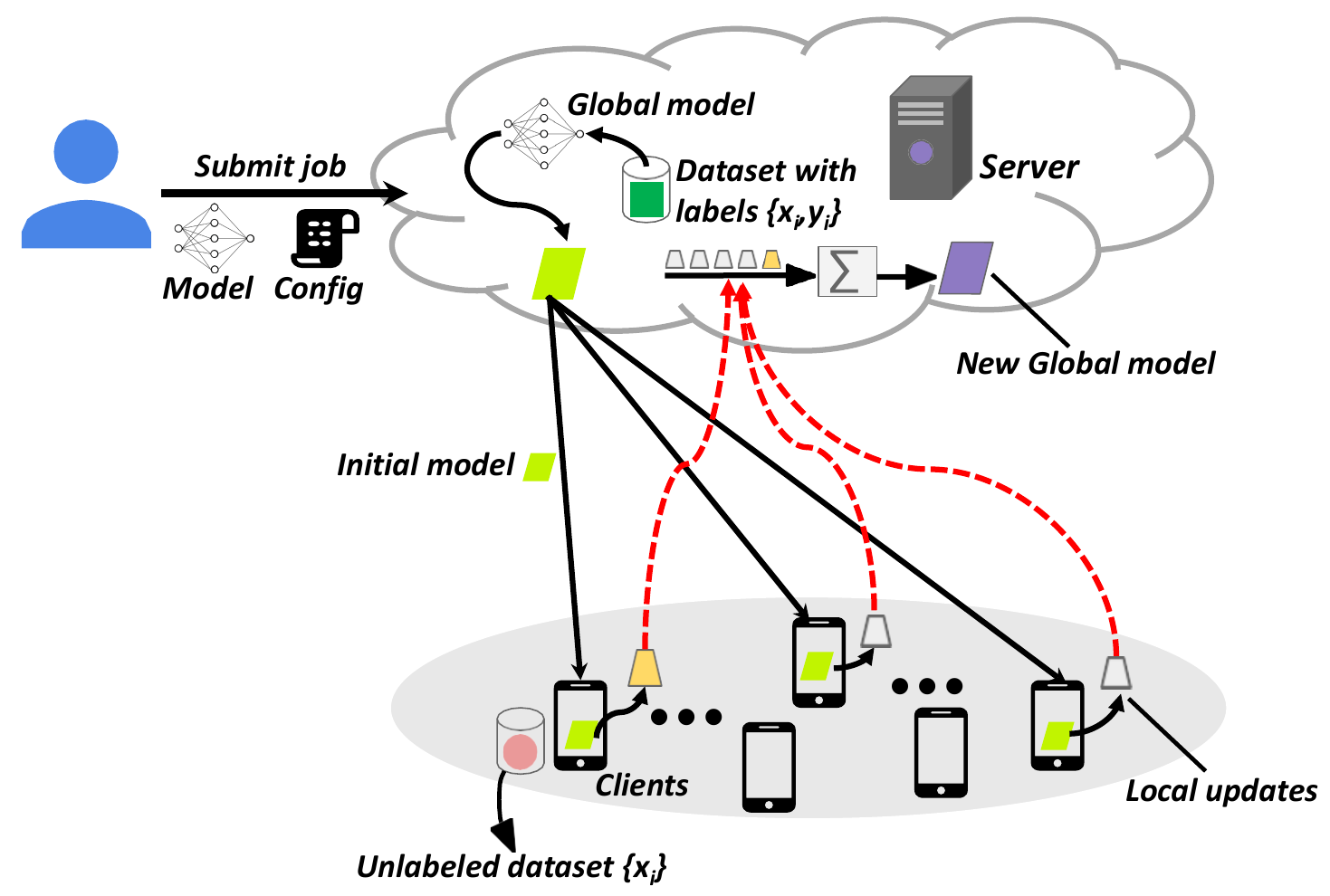}
	\caption{Overview of the semi-supervised federated learning system.}
	\label{fig-2}
	\vspace{-0.5cm}
\end{figure}

\subsection{Semi-supervised Federated Learning}


We first review the limitations of supervised federated learning (SFL). 
%
%
SFL learns the statistical laws of large-scale labeled data (such as statistical distribution and feature space)) to ensure high performance on unseen datasets. However, in many application domains like medical and financial fields, tagged data is precious and difficult to obtain. 
In addition, SFL generally achieves perfect accuracy on the training dataset if trained sufficiently, but it may not generalize well to the test data. The reason is that the diversity of training data is often limited. Among all the existing methods, a good strategy for improving generalization is to train on a larger training dataset \cite{carlini2021poisoning}. Unfortunately, the cost of collecting such large datasets is high. 




To address these issues, we present semi-supervised federated learning (SSFL), which take advantages of semi-supervised learning techniques~\cite{zhai2019s4l,kingma2014semi,wang2007label,yuan2020revisiting,berthelot2019mixmatch,sohn2020fixmatch} to perform training over the data that lacks labels, as illustrated in Fig. \ref{fig-2}. In the SSFL setting, the server $\mathcal{S}$ holds a small labeled dataset ${\mathcal{D}_s} = \{ ({x_i},{y_i})\} _{i = 1}^{{N_s}}$, where $N_s$ denotes the number of samples in the labeled dataset. For $k$ clients in SSFL, each client hold an unlabeled dataset $\mathcal{D}_k$ = $\{({x_i})\}_{i = 1}^{N_k}$, where $N_k$ $({N_k} \gg {N_s})$ denotes the number of samples in the local unlabeled dataset. Similar to a typical FL system \cite{ref-5}, the goal of the server and clients in SSFL is to collaboratively train a global model in high quality. Note that it is challenging for SSFL to achieve this goal, due to the fact that the client has unlabeled data. 
%
%
The steps of an SSFL training method are as follows:

\noindent \textbf{\textit{Phase 1, Initialization:}} The server selects a certain proportion $q$ $(0 < q  \leqslant  1)$ of clients from all clients to participate in an SSFL task (e.g., prediction tasks). Unlike the typical FL, the server not only aggregates the updates uploaded by the clients, but also trains an initialized global model $\omega_0$ on its own labeled dataset $\mathcal{D}_s$. On the server side, the goal is to minimize the following objective function:
\begin{equation}\label{eq-3}
\begin{array}{c}
  \mathop {\min }\limits_\omega  {\mathcal{L}_s}(\omega ), \hfill \\
  {\mathcal{L}_s}(\omega ): = \frac{1}{{{N_s}}}\sum\nolimits_{\{ ({x_i},{y_i})\}  \in {\mathcal{D}_s}} {\ell [{y_i},} f(\phi ({x_i});\omega )], \hfill \\ 
\end{array} 
	\end{equation}
	where ${\mathcal{L}_s}(\omega)$ is the loss function, $\ell ( \cdot , \cdot )$ is the cross-entropy loss, and $\phi ( \cdot )$ is the data augmentation function (e.g., flip-and-shift augmentation \cite{ref-58} and RandAugment \cite{ref-59}). After that, the server broadcasts the initialized global model $\omega_0$ to assign pseudo labels to the selected clients.
	
\noindent \textbf{\textit{Phase 2, Local Training:}} Each client trains the received global model $\omega_0$ on its own unlabeled dataset $\mathcal{D}_k$. Specifically, the client utilizes the global model $\omega_0$ to assign pseudo-label to the unlabeled dataset to train the local model $\omega_0^k$. In the semi-supervised learning, one generally uses pseudo-labels generated by weak data augmentation methods (e.g., flip-and-shift) to train on samples generated by strong data augmentations methods (e.g., shift-and-crop) \cite{ref-60}. The reason is that these data augmentation methods can generate a wide variety of samples, which can improve the performance of semi-supervised learning. Thus, we follow \cite{ref-60} to define the following objective function:
	\begin{equation}\label{eq-4}
	\begin{aligned}
  \mathop {\min }\limits_{{\omega _k}} {\mathcal{L}_k}(\omega ), 
  {\mathcal{L}_k}(\omega )&: = \frac{1}{{{N_k}}}\sum\nolimits_{\{ ({x_i})\}  \in {D_k}} {\gamma (\max ({{\bar y}_i}) \geqslant \lambda )}  \hfill \\
   &\times \ell (\arg \max ({{\bar y}_i}),f(\varphi ({x_i});{\omega _k})), \hfill \\ 
	\end{aligned}
	\end{equation}
where $\lambda $ is the threshold hyperparameter, which helps the model decide which samples have high confidence to be correctly labeled, $\varphi ( \cdot )$ is the strong data augmentation function, $\gamma(\cdot)$ is the indicator function, and ${{\bar y}_i} = {f_k}(\phi ({x_i});\omega )$ is the prediction of the model $f_k$ on the augmented sample generated by weak data augmentation methods $\phi ( \cdot )$. Then all clients upload the updates to the server.

\noindent \textbf{\textit{Phase 3, Aggregation:}} The server uses an aggregation algorithm like FedAVG \cite{ref-5}, i.e., ${\omega _{t + 1}} \leftarrow \sum\limits_{k = 1}^K {{p_k}} \omega _t^k $, to aggregate all the updates to obtain a new global model $\omega_t$. After that, the server continues to train the global model on the dataset $D_s$.

\noindent \textbf{Remark:} For Eq. \eqref{eq-3} and Eq. \eqref{eq-4}, one may use Stochastic Gradient Descent (SGD) with momentum over a mini-batch to optimize. For aggregation algorithms, one may use other aggregation algorithms depending on the specific application scenario. The above steps are repeated until the global model converges.



\begin{figure*}[!t]
	\centering
	\includegraphics[width=0.80\linewidth]{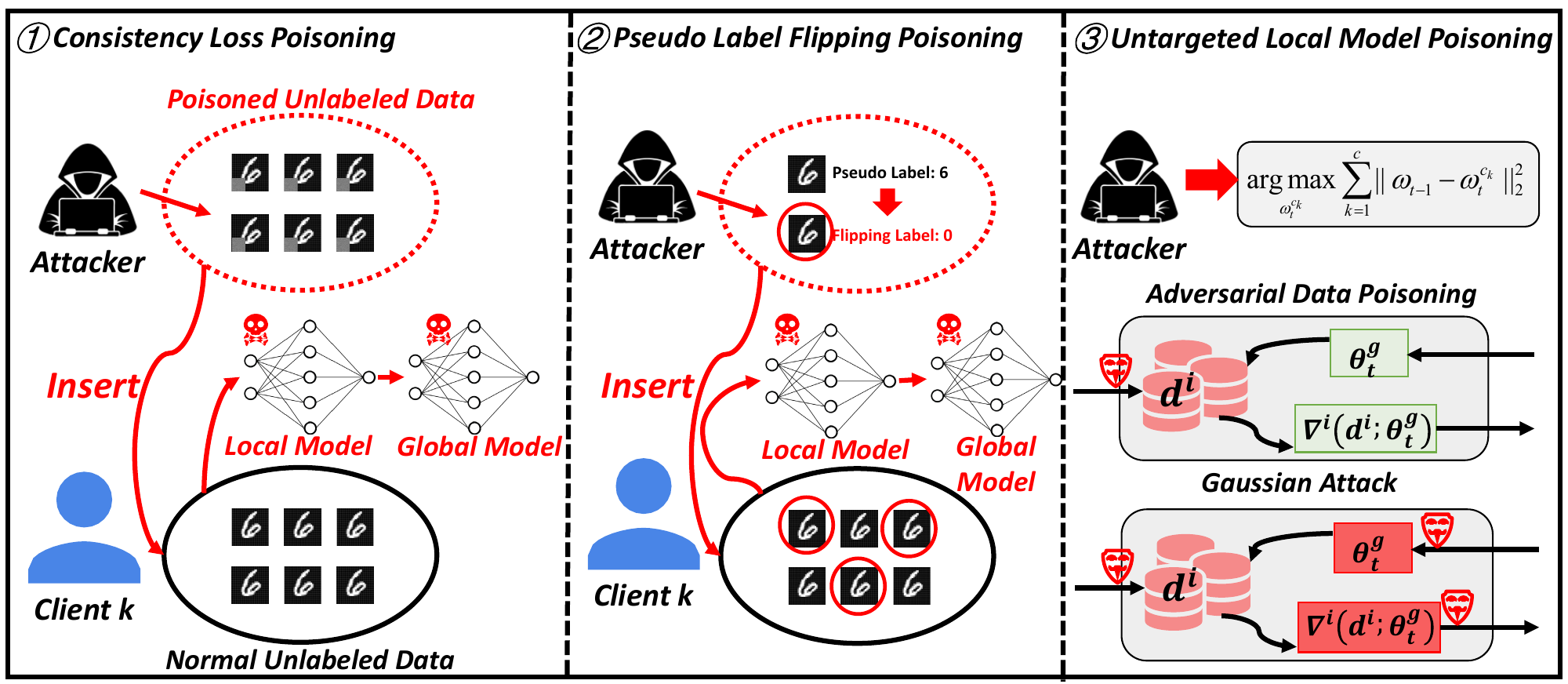}
	\caption{Our proposed poisoning attacks in SSFL.}
	\label{fig}
\end{figure*}

\subsection{Attack Model}
In this paper, we consider an adversary who controls some clients to train local machine learning models on unlabeled datasets and upload the poisoned local model updates to the server for aggregation. To poison the global model, we assume that the adversary has the following abilities: (1) he can control a group of compromised clients with an index length of $c$; (2) he can manipulate the unlabeled dataset; (3) he can access the local training samples $\{x_i\}$ of the client $k$ to generate the corresponding poisoned samples $\mathcal{D}_k^p$. Specifically, the poisoning adversary $\mathcal{A}$ constructs a set of poisoning examples on the unlabeled dataset, and the formal definition is as follows:
\begin{equation}
{\mathcal{D}_k^p} \leftarrow \mathcal{A}({x^*},{y^*},N_p,{f_k},{T_k},\mathcal{D}_k';{c_k}),
\end{equation}
where $x^* \in \mathcal{D}_k'$ is the input to be poisoned by the adversary on client $c_k$, ${y^*} \ne f({x^*})$ is the desired incorrect label or target label, $N_p$ is the number of samples that can be injected, $f_k$ is the local machine learning model on client $c_k$, $T_k$ is the local training algorithm on client $c_k$, and a subset of unlabeled samples $\mathcal{D}_k' \subset \mathcal{D}_k$.

\noindent \textbf{Attack Goal:} 
The adversary's goal is to make the classifier learned at the server (i.e., global model $f_\omega$) misclassify the target samples, i.e., ${f_\omega}({x^*}) = {y^*}$ or ${f_\omega}({x^*}) \ne  {y}$. In particular, we require that the adversary can only poison a small part of the unlabeled dataset (i.e., the poisoning rate $p$), that is, $|{\mathcal{D}_k^p}| \leqslant p \cdot |\mathcal{D}_k|$. More specifically, we expect $p$ to be less than $1\%$ (i.e., $p \leqslant 1\% $). 





\section{The Proposed Attacks} \label{sec-3}

\subsection{Design Intuition}

Recall that semi-supervised learning utilizes the self-supervised nature to achieve high performance on unlabeled datasets. Simply put, SSFL allows the local client to use the received global model to assign pseudo-labels on the unlabeled dataset and expect that it will self-learn to solve the learning obstacles caused by the unlabeled data.
As, semi-supervised learning relies heavily on the correctness of label information,  especially the label information of unlabeled data. We observe that this requirement creates an new opportunity for adversaries to poison the model. Following this natural characteristic behind semi-supervised machine learning, we design poisoning attacks against SSFL, as shown in Fig. \ref{fig}. In particular, our attack is generic and not specific to any particular algorithms. 

In this paper, we propose two attack strategies to demonstrate the vulnerability of SSFL to the poisoning attacks. Specifically, we focus on the poisoning attacks that directly damage the performance of the SSFL system. To this end, we consider that the adversary poisons the system by modifying the pseudo-label of the local dataset (i.e., data poisoning attacks) or the local model updates of the client (i.e., model poisoning attacks). In this context, we design two attack strategies to poison semi-supervised federated learning systems by inserting or manipulating poisoned unlabeled datasets. Specifically, we exploit the unlabeled dataset to propose the following three poisoning attacks: consistency loss poisoning, pseudo label flipping poisoning, and untargeted local model poisoning.


\subsection{Consistency Loss Poisoning}
In most cases, the client utilizes the received global model to assign pseudo-labels for unlabeled data and uses the consistency loss to train the local model. Specifically, this consistency loss includes fully-supervised and semi-supervised losses, whereas training semi-supervised loss relies on pseudo-labels objective of unlabeled data. For example, in \cite{ref-38}, Jeong \textit{et al.} proposed inter-client consistency loss to train SSFL. The insight of these consistency loss methods is to transfer the label of a labeled sample to nearly unlabeled samples with high confidence. While many different consistency loss methods have been proposed, they all help the model generate a decision boundary by utilizing correct label information of unlabeled data to achieve high-accuracy classification in the feature space. It indicates that if we can poison the consistency loss, this decision boundary is disturbed so that the infected model gets incorrect classification results. 
%
 
Following the above intuition, we design a poisoning attack against SSFL. The idea is similar to \cite{carlini2021poisoning}; that is, we insert some poisoned data samples in the unlabeled data part of the training data. The main difference from \cite{carlini2021poisoning} is that we perturb to poison the data instead of interpolation, and we implement it in a distributed setting, which is much more difficult than implementing it in a centralized environment due to the non-i.i.d.. setting and learning paradigm. Specifically, we follow the classic assumption that the adversary understands (at least part of) the knowledge of the training data. The adversary needs to use the inserted poisoning data on the given target image $x'$ (selected from the labeled dataset) to complete the misclassification of the specified target, i.e., $f_k(x')=y^*$. Thus, for the client $c_k$, the inserted $N_p$ samples can be expressed as follows:
\begin{equation}
\{ {x_\epsilon }\} _{i = 0}^{N_p - 1} = P(x',{x^*},\epsilon ),
\end{equation}
where $N_p$ is the number of poisoned unlabeled data, $\epsilon$ is a slight disturbance, and $P( \cdot , \cdot , \cdot )$ is the poisoning function. Note that we use Gaussian noise to generate a poisoned sample; thus, with ${\sigma ^2} \in [0,1]$ denoting the variance of Gaussian noise, we have:
\begin{equation}
P(x',{x^*},\epsilon  ) = \left\{ \begin{gathered}
  x',\epsilon =\sigma^2  = 0 \hfill \\
  {x^*},\epsilon =\sigma^2  \ne 0 \hfill \\ 
\end{gathered}  \right..
\end{equation}

Considering that such a poisoning strategy is to poison unlabeled samples near the labeled data, then we have: $||x' - x^*||_p \leqslant \epsilon$. Furthermore, the consistency loss in SSFL includes semi-supervised loss, so we can optimize the following function to poison this consistency loss:
\begin{equation}
\begin{aligned}
  {\ell _p}( x ) = &\mathop {\min }\limits_{{x^*}} \sum\limits_{i = 0}^{{N_p} - 1} {KL[} p_\omega ^*(y|{x^*})||{p_\omega }(y|x')] \hfill \\
  &s.t.{\text{ }}||x' - {x^*}|{|_p} \leqslant \epsilon  \hfill \\ 
\end{aligned} ,
\end{equation}
where ${\ell _p}( x )$ is the poisoning semi-supervised training loss, $KL$ represents the Kullback-Leibler divergence (a.k.a. relative entropy), $p_\omega(\cdot)$ is the output of the local model, and $\epsilon$ is is a small constant representing a small perturbation. As a result, with more and more poisoned unlabeled data, the classification results of the model are moving in the direction of misclassification.

When we utilize such a poisoning strategy to poison unlabeled data, the model's training process is diverted along the wrong path. The reason is that once the model sets the poisoned unlabeled data as the target label expected by the adversary in a high confidence manner, the adversary can dynamically adjust the disturbance $\epsilon $ along this path so that the model continues to mistakenly believe that these pseudo-labels of the unlabeled data are correct.

\noindent \textbf{Remark:} We emphasize that the designed attack uses the nature of semi-supervised learning to infect the model, which is a stealthy poisoning attack. As the semi-supervised training progresses, unlabeled samples near the labeled samples are assigned the same label $y$ as the poisoned unlabeled data. The adversary just needs to repeat this process and continue to poison more unlabeled data along this path. In the end, all injected examples $x^*$ will be labeled $y$ in the same way. This means that in the end, $f({x_\epsilon  }) = f({x^*}) = {y^*}$ will also complete the poisoning attack.

\subsection{Pseudo Label Flipping Poisoning} 
Inspired by label flipping attacks in SFL, we design a label flipping poisoning attack that flips only pseudo-labels. In this attack, we assume that the adversary can manipulate the client's local dataset $\mathcal{D}_k = \{ {x_i},{{\hat y}_i}\} _{i = 1}^{N_k}$ (a dataset with pseudo-labels) to implement a label-flipping poisoning. To be specific, the adversary flips the pseudo label in the $\mathcal{D}_k^p$ to a specified target label $y^*$ to fool the model, thereby harming the performance of the model. Formally, the training loss function of this attack for client $k$ is defined as follows:
\begin{equation}
{\ell _l}(x) = \mathop {\arg \min }\limits_{{x} \in \mathcal{D}_k} (\sum\limits_{i = 1}^{N - {N_p}} {\ell [{f_k}({x_i}),{{\hat y}_i}] - \sum\limits_{j = 1}^{{N_p}} {\ell [{f_k}({x_j}),y_j^*]} ),} 
\end{equation}
where $\ell _l(\cdot,\cdot) $ is the local loss function, $N_p$ is the number of poisoning samples, and $\hat y$ is the pseudo-label. 

\noindent \textbf{Remark:} Although the methodology of the above poisoning attack is similar to the label flipping attack in SFL, there are several differences. First, the goal of label flipping attacks in SSFL is to degrade the learner's performance on the server to label pseudo-labels so that the pseudo-labels cannot be set correctly. 
%
%
In contrast, label flipping attacks in SFL poisoning the local data. Second, the label flipping attack in SSFL can gradually become stronger as the attack progresses. The reason is that SSFL relies heavily on the correctness of pseudo-label information.

\subsection{Untargeted Local Model Poisoning}
In addition to the above data poisoning attacks, we can also utilize unlabeled data to craft poisoned model updates to poison the global model. Here, we propose an untargeted local model poisoning attack by setting wrong pseudo labels. Our insight is that the current local model update, which is significantly different from the previous round of global model updates, is not conducive to the training of the global model. In this attack, the adversary manipulates a subset (index length is $c$) of the client set to send well-designed poisoning model updates to the server to destroy the performance of the global model. Let the poisoned model updates of the compromised client $k$ be $\omega _t^{{c_k}}$. The objective function of this attack is defined as follows:
\begin{equation}\label{eq-8}
{\ell _m}(\omega ) =  \mathop {\arg\max }\limits_{\omega _t^{{c_k}}} \sum\limits_{k = 1}^c {||{\omega _{t - 1}}}  - \omega _t^{{c_k}}||_2^2.
\end{equation}

\noindent \textbf{Remark:} For simplicity and demonstration, we introduce two attack methods to optimize Eq.\eqref{eq-8}. The first method is to poison the global model by making the local model updates follow a Gaussian distribution. The second one is that the adversary crafts an adversarial dataset $\mathcal{D}_{adv}$ to maximize the difference between the gradient descent directions produced by the client and those on the server.



\subsection{Attack Scenarios}
We envision two attack scenarios: a centralized attack scenario and a distributed attack scenario~\cite{xie2019dba}. In \cite{xie2019dba}, Xie \textit{et al.} explored a centralized attack scenario and a distributed attack scenario in FL for backdoor attack. For the former, the adversary can only control unlabeled data with a proportion of $(c*p\%)$ poisoned by a compromised client. In the latter scenario, the adversary can control $c$ compromised clients to poison $(c*p\%)$ unlabeled data together. Testing under these two attack scenarios can comprehensively evaluate the effectiveness of our proposed attack.

\noindent \textbf{Centralized Poisoning Scenario.} In this scenario, the adversary can only control one compromised client, but it can poison more local unlabeled data (i.e., $(c*p\%)$) of the entire of the local dataset). In fact, such an attack is more practical because it is very challenging to control multiple compromised clients in the real world.


\noindent \textbf{Distributed Poisoning Scenario.} In this case, the adversary can control $c$ compromised clients to poison their local unlabeled dataset with a total proportion of $p\%$. For a fair comparison with the centralized attack scenario, we allow these compromised clients to poison only as much unlabeled data as in the centralized scenario. In a sense, such an attack scenario can represent an upper bound on the attack effect of the proposed attack. The reason is that the adversary controls enough compromised clients in this scenario.



\section{Defense}\label{sec-4}
In this section, we present a defense to address the proposed poisoning attacks. As seen from the above, ``correct'' and credible label information and aggregated high-quality model updates are the keys to SSFL's ability to resist poisoning attacks. 
We believe that techniques that identify potential malicious clients can provide high-quality label information and model updates, so as to mitigate these attacks.

\subsection{Design Intuition}
In Sec. \ref{sec-3}, we have thoroughly investigated the proposed poisoning attacks in SSFL. We design a minimax optimization-based client selection strategy that can select clients who hold ``correct'' label information and high-quality model updates. Next, we first illustrate the intuition behind our defense.

\begin{figure}[!t]
	\centering
	\includegraphics[width=1\linewidth]{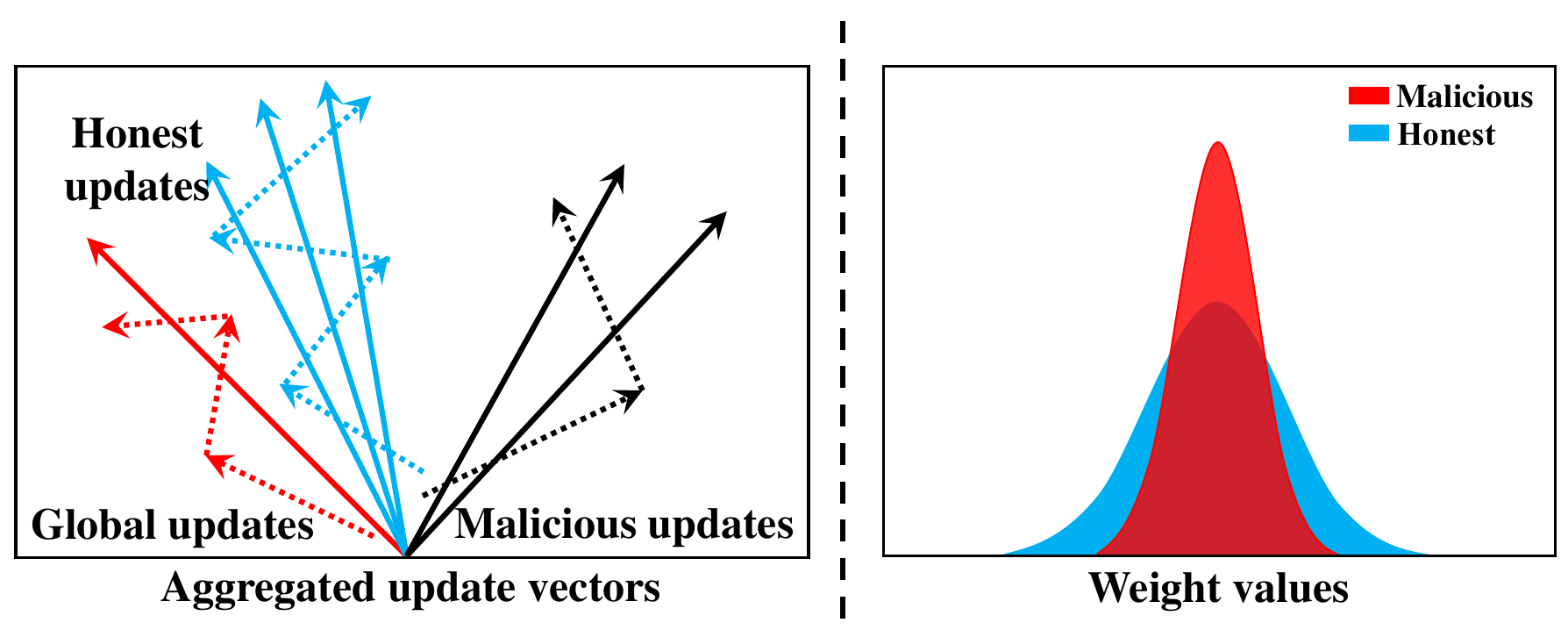}
	\caption{Comparison of the client's aggregated update vector with the update direction of the global model \textit{\textbf{(left)}}. Comparison of the weight update distribution of honest participants and poisoned clients \textit{\textbf{(right)}}.}
	\label{fig-3}
\end{figure}

As illustrated in Fig. \ref{fig-3}, we explore the difference between the aggregated updates and weight update distribution between poisoned clients and honest participants. First, in Fig. \ref{fig-3} \textbf{\textit{(left)}}, we find that the update direction of the an honest client is close to the update direction of the server. On the contrary, the update direction of the aggregated update vectors of the malicious clients is different from the update direction of an honest participant. Second, in Fig. \ref{fig-3} \textbf{\textit{(right)}}, the distribution of updates generated by malicious clients is obviously different from the distribution of updates generated by honest participants. 

\subsection{Minimax Optimization-based Client Selection Strategy}


Motivated by the above observations, we propose to use the distribution difference (direction difference) between the client's weight update (aggregated update vectors) and the server's weight (aggregated update vectors) as an indicator to measure the update quality of the client updates. In such a way, the server only aggregates model updates uploaded by clients with high-quality updates to obtain the high performing global model. Furthermore, such a strategy allows the clients to manually check the correctness of the label information of the unlabeled samples, which can significantly reduce the possibility of the adversary injecting the poisoned label information. Thus, the formal definition of the minimax optimization strategy is as follows:
\begin{equation}\label{eq-5}
\left\{ \begin{array}{l}
	\mathop {\min }\limits_\omega  {\lambda _{{\ell _2}}}||{\omega _t} - \omega _t^k||_2^2 + {\lambda _{{\ell_1}}}||\omega _t^k{||_1},\\
	\max \sum\limits_{k = 1}^K {({\Delta _{k,t}}|{\Delta _{k,t}} \ge \beta  ),} 
\end{array} \right.
\end{equation}
where ${\lambda _{{\ell_2}}}$ and ${\lambda _{{\ell_1}}}$ are the relative weights of $\ell_2$ and $\ell_1$ regularization, respectively, $\omega_t$ represents the global model's parameters, $\omega_t^k$ represents the $k$-th client's local model parameters, ${\Delta _{i,t}}$ represents a measure of the update quality of the $t$-th round of training, and $\beta$ is the threshold to help the server to decide which updates have high confidence to be aggregated. The first item of Eq. \eqref{eq-5} means that we should select those clients whose update direction is close to the global model update direction as much as possible. The second term of Eq. \eqref{eq-5} means that we choose as many clients as possible whose distance or distribution is close to the global model.

In this paper, we use the cosine similarity and Wasserstein distance \cite{ref-61} with the global model $\omega ^*$ to indicate the update quality of the client. The formal definition of the cosine similarity measure is as follows:
\begin{equation}\label{eq-6}
\scriptsize
({\Delta _{k,t}}|{\Delta _{k,t}} \geqslant \beta ) = (\cos (\sum\limits_{t = 1}^T {{\mathbf{w}}_t^*} ,\sum\limits_{t = 1}^T {{\mathbf{w}}_t^k} )|\cos (\sum\limits_{t = 1}^T {{\mathbf{w}}_t^*} ,\sum\limits_{t = 1}^T {{\mathbf{w}}_t^k} ) \geqslant \beta),
\end{equation}
where $\sum\limits_{t = 1}^T {{\mathbf{w}}_t^k} $ is the sum of historical updates of the $k$-th client training. Note that the difference between the definition of cosine similarity we designed and the typical cosine similarity is that we use the sum of the client's historical updates. The reason is that the sum of historical updates can better reflect the reputation of the client \cite{ref-78}. In this context, the client's reputation is used to measure the historical behavior of this client.

If we regard the client's update as a distribution, then the formal definition of the method using Wasserstein distance as the metric is as follows:
\begin{equation}\label{eq-7}
{\Delta _{k,t}} =  - {W_d}({\mathbf{w}}_t^*,{\mathbf{w}}_t^k) =  - \mathop {\inf }\limits_{\gamma {\text{ }} \sim \;\prod {({\mathbf{w}}_t^*,{{\mathbf{w}}_{k,t}})} } {{\text{E}}_{(x,y) \sim \gamma }}[||x - y||],
\end{equation}
where $\gamma  \sim \;\prod {({\mathbf{w}}_t^*,{\mathbf{w}}_t^k)}$ is  the joint distribution of ${\bf{w}}_t^*$ and ${\mathbf{w}}_t^k,{{\text{E}}_{(x,y) \sim \;\gamma }}[||x - y||]$ represents the expected value of the distance of the sample pair $(x, y)$ in the joint distribution $\gamma$, and $W_d( \cdot , \cdot )$ is the lower bound of the expected value of the Wasserstein distance from the sample. Specifically, we can use the distance between the distribution of the client's local updates and the distribution of the global model to select clients with high-quality updates. In particular, we assume ${\bf{w}}_t^* \sim \mathcal{N}({\mu _t},{\Sigma _t})$ and ${\mathbf{w}}_t^k \sim \mathcal{N}(\mu _t^k,\Sigma _t^k)$, then the $2$-Wasserstein metric is given by:
\begin{equation}\label{eq-12}
\begin{aligned}
  {\Delta _{k,t}}^G &=  - {W_d}_2^2({\mathbf{w}}_t^*,{\mathbf{w}}_t^k) = ||{\mu _t} - \mu _t^k||_2^2 \hfill \\
  & + tr({\Sigma _t} + \Sigma _t^k - 2{(\Sigma _t^{\frac{1}{2}}\Sigma _t^k\Sigma _t^{\frac{1}{2}})^{\frac{1}{2}}}) \hfill \\ 
\end{aligned},
\end{equation}
where ${\mu _t}$ ($\sum_t$) and $\mu _t^k$ ($\sum_t^k$) are the mean values (variances) of the global weights' distribution and the sum of client history updates' distribution, respectively.

\subsection{Detailed Algorithm}
\noindent \textbf{Quality-based Aggregation.} Since not all local models learned on pseudo-labeled data can be as reliable as models learned on labeled data, evaluating the quality of the locally learned knowledge is crucial. To this end, we propose a new \textit{quality-based aggregation rule $Q( \cdot )$} instead of FedAvg to aggregate model updates. Specifically, we utilize a server-side test set to calculate the quality score (i.e., accuracy) of each client's local model, thereby enhancing the effect of high-quality models while minimizing the negative effects of low-quality models. Next, we give the definition of \textit{quality-based aggregation rule $Q( \cdot )$} as follows:

\begin{definition}\label{defi-2}
\textbf{(Quality-based Aggregation Rule $Q( \cdot )$) (QAR):} For the $t$-th training round, we assume that $\mu  _t^i$ denotes the quality score of $k$-th local model $\omega _t^k$ on the server-side test set and $\tau   = \sum\limits_{k = 1}^K {\mu  _t^i} $ denotes the sum of all quality scores on $k$ clients, thus, the $Q(\omega ) = \sum\limits_{k = 1}^K {\frac{{\mu _t^k}}{\tau }} \omega _t^k.$ Referring to FedAvg, the formal definition of quality-based aggregation rule is as follows:
\begin{equation}\label{eq-11}
{\omega _{t + 1}} = {\omega _t} + \frac{1}{K}Q({\omega}).
\end{equation}
\end{definition}

Based on the above-proposed defense strategy, quality-based aggregation rule, and the SSFL training procedures, our entire SSFL secure training is summarized in Algorithm \ref{al-1}. Thus, we describe the algorithm step by step as follows:
\begin{itemize}
    \item \textit{\textbf{Step 0 Initialization:}} First, the server selects a subset of clients set to participate in an SSFL training. Second, unlike the typical FL, the server not only aggregates the updates uploaded by the clients, but also trains an initialized global model $\omega_t$ on its own labeled dataset $\mathcal{D}_s$.
    \item \textit{\textbf{Step 1 Pseudo-labeling:}} Each client sets the threshold hyperparameters $\lambda = 0.95$ to set pseudo label. Note that $\lambda$ is to set the confidence of the pseudo-label which affects the semi-supervised performance. We follow \cite{berthelot2019mixmatch} to set $\lambda = 0.95$. Then the clients use the received global model $\omega_{t}$ to label the local unlabeled dataset. Furthermore, they use data augmentation methods to generate diverse labeled data.
    \item \textit{\textbf{Step 2 Local Training:}} Each client trains the received global model $\omega_t$ on its pseudo-labeled dataset $\mathcal{D}_k$ and uploads its local updates.
    \item \textit{\textbf{Step 3 $\mathrm{QRA}$ Aggregation:}} The server utilizes minimax optimization strategy to select clients with high-quality updates by using Eq. \eqref{eq-5}--Eq. \eqref{eq-7}. Finally, the server uses $\mathrm{QRA}$ to aggregate the selected updates, i.e., ${\omega _{t + 1}} = {\omega _t^k} + \frac{1}{K}Q({\omega})$ (refer to Defi. \ref{defi-2}).
\end{itemize}

\begin{algorithm}[t!]
	\caption{Minimax Optimization-based Client Selection Strategy.}\label{al-1}
	\LinesNumbered 
	\KwIn{The labeled dataset $\mathcal{D}_s$ on the server, the unlabeled dataset $\mathcal{D}_k$ on the client $k$, the threshold hyperparameters $\alpha$ and $\beta$, and the cross-entropy loss is $\ell ( \cdot , \cdot )$}
	\KwOut{Optimal global model $\omega^*$.}
	\textbf{Server:}\\
	Initialize the global model\\
	\ForEach{training round $t = 1,2,\ldots,T$}{
		Train the global model $\omega_{t}$ on the labeled dataset $\mathcal{D}_s$ by using $\ell ( \cdot , \cdot )$ (Refer to Eq. \eqref{eq-3})\;
		Broadcast the global model $\omega_{t}$ to all the clients\;
	\textbf{Client:}\\
	\ForEach{client $k$, $k \in \{1,2,\ldots,K\}$}{
		Receive global model $\omega_{t}$\;
		Use the received global model $\omega_{t}$ to label the local unlabeled dataset\;
		Use data augmentation methods to generate diverse labeled data\;
		Train the local model $\omega_{t}^k$ on the local pseudo-labeled dataset $\mathcal{D}_k$ by using Eq. \eqref{eq-4}\;
		Upload the local updates to the server\;
	}
	\textbf{Server:}\\
		Utilize minimax optimization strategy to select clients with high-quality updates by using Eq. \eqref{eq-5}--Eq. \eqref{eq-12}\;
		Use $\mathrm{QRA}$ to aggregate the selected updates, i.e., ${\omega _{t + 1}} = {\omega _t^k} + \frac{1}{K}Q({\omega})$ (refer to Defi. \ref{defi-2})\;
	}
	\Return $\omega^* $.
\end{algorithm}

\section{Experiments}\label{sec-5}
\subsection{Experiment Setup}\label{sec-5-1}
In this section, we evaluate our designed poisoning attacks on MNIST and CIFAR-10 datasets. All experiments are implemented on the same computing environment (Linux Ubuntu 18.04, Intel i5-4210M CPU, 16GB RAM, and 512GB SSD) with Pytorch and PySyft \cite{ref-70}.

\noindent \textbf{Models:} In this experiment, we use a simple deep learning model (i.e., CNN with 2 convolutional layers followed by 1 fully connected layer) for classification tasks on the MNIST dataset and use the AlexNet \cite{ref-71} model for classification tasks on the CIFAR-10 dataset.

\noindent \textbf{Datasets:} MNIST is a handwritten digital image dataset that contains 60,000 training samples and 10,000 test samples. Each image consists of $28 \times 28$ pixels, and the label is one-hot code: 0--9. The CIFAR-10 dataset consists of 10 types of $32 \times 32$ color images, with each type containing 6,000 images for 60,000 images. Among them, 50,000 images are used as the training dataset, and 10,000 images are used as the test set. The pixel values are normalized into $[0,1]$, and the label is one-hot code: 0--9. Note that in our SSFL system, we remove the label information of the local dataset on the client-side.

\noindent \textbf{datasettings:} For data assignment procedures, we assign $N_s$ labeled training samples to the server, and the remaining unlabeled training samples are assigned to $K$ clients. For the i.i.d.. data distribution setting, we assign the unlabeled data of $d=10$ classes to each client. For the non-i.i.d.. data distribution setting, we evenly distribute the unlabeled data to each client while keeping \textit{only two categories} of unlabeled data for each client. Note that the classes of each client in the non-i.i.d. setting are randomly assigned. Furthermore, we allow the client to make precise pseudo-label settings in the first few training rounds.

\noindent \textbf{Attack Settings:} We assume that the adversary manipulates $c$ compromised clients to launch three types of the proposed poisoning attacks: consistency loss poisoning, pseudo label flipping poisoning, and untargeted local model poisoning.
For the pseudo label-flipping attack, the adversary modifies one class of pseudo label of the local dataset to another class. Specifically, for the MNIST dataset, the adversary changes all the training samples with the pseudo label ``1'' to the pseudo label ``7''; for the CIFAR-10 dataset, the adversary changes all the training samples pseudo labels with pseudo label ``dog'' to ``cat''. We use the terms ``Gaussian attack'' and ``adversarial data poisoning'' to denote the Gaussian noise attack method and the adversarial data attack method in untargeted local model poisoning, respectively. For the Gaussian attack, the adversary controls compromised clients to upload its update $\omega _t^k$ from a Gaussian distribution with mean $\frac{1}{{K - c}}\sum\nolimits_{\omega  \notin {\omega _c}} {\omega _t^k} $ and variance 10. For the adversarial data poisoning attack, the adversary can access compromised clients' local training dataset to generate an adversarial poisoning dataset $\mathcal{D}_{adv}$ to poisoning the global model. 


\noindent \textbf{Parameters:} We set the number of clients $K=100$, the number of compromised clients $c=30$, the proportion of client participation $q=1$, training round $T=250$, local training epoch $E=5$, learning rate $\eta  = 0.001$, mini-batch size $B=32$, and the number of training samples $N_s=10,000$ on the server-side. Furthermore, we use SGD with momentum over a mini-batch to optimize our model.

\noindent \textbf{Evaluation Metric.} We use the Attack Success Rate (ASR) to evaluate the attack performance, which is defined as follows:
\begin{definition}
\textbf{Attack Success Rate (ASR).} The adversary chooses a target class $y^*$, and defines the ASR as the fraction of correctly classified inputs that are not labeled as the target class but misclassified to the target class after the poisoning unlebeled samples are injected:
\begin{equation}
\psi (f({x^*},{\omega ^*}) = {y^*}|y \ne {y^*},f(x,{\omega ^*}) = y).
\end{equation}
\end{definition}
Furthermore, we denote by $Acc_{test}^*$ the maximum accuracy that the global model converges without any attack. We use $Acc_{test}^p$ to denote the maximum accuracy the global model can achieve under a given untargeted poisoning attack. Here, we define attack impact $\Delta $ as the reduction in global model accuracy due to the attack, i.e., 
\begin{equation}
 \Delta  = Acc_{test}^* - Acc_{test}^p.  
\end{equation}

\noindent \textbf{Semi-Supervised Learning Methods:} We apply two of the most popular semi-supervised learning techniques (i.e., FixMatch \cite{sohn2020fixmatch} and MixMatch \cite{berthelot2019mixmatch}) to FL to build an SSFL. In addition, we also explore the impact of different semi-supervised learning methods on the attacks we designed.

\subsection{Evaluation}
Our evaluation goal has three-fold: \textit{(i)} To show the poisoning performance of the proposed poisoning attacks, we report the ASR and attack impact $\Delta $ of the proposed attacks; \textit{(ii)} To explore the parameter sensitivity of the proposed algorithm, we conduct extensive case studies around the parameter $p$ and SSFL methods; \textit{(iii)} To verify the effectiveness of the proposed defense, we evaluate it in terms of accuracy and ASR under the proposed poisoning attacks.

\subsubsection{Poisoning Performance} In this section, we conduct a series of experiments on the MNIST and CIFAR-10 datasets to demonstrate the vulnerability of SSFL to the above-mentioned poisoning attacks.

\noindent \textbf{Poisoning Performance with Different $p$.}
We evaluate the poisoning performance of the designed poisoning attack under the i.i.d.. datasetting. Note that in this experiment, we use FixMatch as our semi-supervised technique for SSFL. Please refer to the following for the experiment of different semi-supervised methods on poisoning performance. In this experiment, we set different poisoning rate $p \in \{ 0.1 ,0.2 ,0.5 ,1\} $ to evaluate our attack. Specifically, for targeted poisoning attacks (i.e., consistency loss poisoning), we select 100 poisoned samples as input evaluate the model and report ASR. Furthermore, to comprehensively verify the performance of the designed attack, we consider two attack scenarios: \textbf{centralized poisoning and distributed poisoning scenarios}. The former indicates a single compromised client with a $c*p\%$ ratio, and the latter indicates all compromised clients of poisoned unlabeled data together to poison unlabeled data with a ratio of $c*p\%$. In both scenarios, the ratio of poisoned data is the same; however, it can be distributed to a single client (centralized) or multiple clients (distributed).

We first evaluate the poisoning performance of consistency loss poisoning. In a centralized poisoning scenario, as shown in Table \ref{tab-1}, when $p=1\%$, the ASR reaches 75\% on the MNIST dataset, which means that the global model is infected and outputs a lot of misclassification results in the classification task. This experimental result indicates that our proposed consistency loss poisoning successfully poisons the global model in a realistic setting. We also observe from Table \ref{tab-1} that the ASRs of the distributed poisoning scenario on the MNIST and CIFAR-10 datasets are higher than the centralized poisoning scenario, which means that the performance of the attack increases due to the decrease in the number of poisoned samples and the number of compromised clients.

We then evaluate the poisoning performance of the pseudo-label flipping attack and the untargeted local model poisoning attack. According to the experimental results, we find that the above three attack methods cannot successfully poison the global model when the poisoning ratio is $p=\{0.1, 0.2, 0.5\}$. Therefore, we only show the attack performance when this experiment's $p=1\%$ distributed poisoning scenario. Table \ref{tab-6} reports the performance of SSFL subjected to the above-mentioned attacks. From the experimental results, we find that these poisoning attacks have also caused damage to the performance of SSFL, especially label flip attacks. The reason is that semi-supervised technology relies heavily on label information. Furthermore, we can confirm that the attack performance of the consistency loss poisoning attack tailored for SSFL is much stronger than the above three attack methods. The reason is that the consistency loss poisoning attack disrupts the process of semi-supervised learning. On the contrary, the poisoning attack is commonly seen in SFL. Although it also caused damage to SSFL, the power needs to be improved. Therefore, in the following experiments, we focus on exploring the performance of the consistency loss poisoning attack.

\noindent \textbf{Discussion:} We validate our attacks under two different poisoning scenarios. The experimental results show that with our proposed consistency loss poisoning attack, the adversary just needs to manipulate $0.1\%$ of the unlabeled dataset to achieve an attack success rate of at least $80\%$. Under the same circumstances, in SFL, the adversary generally needs to manipulate $1\%  \sim 5\% $ \cite{carlini2021poisoning} of the data to achieve the same goal. This implies that SSFL is much more vulnerable to poisoning attacks than SFL when dealing with poisoning attacks, especially those caused by unlabeled data.

\noindent \textbf{Poisoning Performance with different SSL Methods.}
In this experiment, we investigate the impact of different semi-supervised methods on the consistency loss poisoning attack we designed. Specifically, we use the two most popular methods in the semi-supervised learning community, FixMacth, and MixMatch, as our research objects. In the distributed poisoning scenario, we set $p \in \{ 0.1\% ,0.2\% ,0.5\% ,1\% \} $ and evaluate the infection of the model on the MNIST and CIFAR-10 datasets. As shown in Table \ref{tab-3}, in most cases, the MixMatch method is more infection resistant under the poisoning attack than the FixMatch method. Our analysis shows that the cross-entropy strategy and the strong data augmentation strategy in the MixMatch method have a certain mitigation effect on the attacks. Specifically, these two strategies make it difficult for our attacks to poison unlabeled data near labeled samples, which prevents attackers from using semi-supervised mechanism to poison unlabeled data. However, in FixMatch, it uses consistency loss and weak data augmentation strategies, which are vulnerable to attacks. Nevertheless, when $p=0.1\%$, our attack has a success rate of about 70\% in both methods. Note that the above results may reveal an intuition that the more advanced semi-supervised learning methods are more vulnerable. We guess that the more advanced the SSFL method, the more heavily dependent on the quality of pseudo labels.

\noindent \textbf{Poisoning Performance under Non-i.i.d. Setting.}
In our SSFL setting, the distribution of data across devices is may be different, resulting in a non-i.i.d. setting, which is rarely addressed in FL. Therefore, it is necessary to explore the performance of our poisoning attack in such a setting. Our starting point is that non-i.i.d. data may facilitate the implementation of poisoning attacks and further increase the difficulty of defending against malicious attacks. Similarly, we set $p \in \{ 0.1\% ,0.2\% ,0.5\% ,1\% \} $ in the distributed poisoning scenario to investigate the poisoning performance of the designed attack. It can be seen from Table \ref{tab-5} that non-i.i.d. data does have a positive impact on our attack. For example, when $p=0.1\%$, ASR is higher than the corresponding result in Table \ref{tab-1}. The reason is that the data distribution under the non-i.i.d. setting is relatively concentrated, making it easier for our attacks to interfere with the model classification decision boundary, thereby infecting the model with a higher success rate.

\begin{table}[!t]
	\caption{Poisoning Performance with different $p$ under centralized and distributed poisoning scenarios.}
	\label{tab-1}
		\centering
	\begin{tabular}{ccccc}
		\toprule
		Dataset& $p\%$  & ASR (Centralized)&ASR (Distributed)\\
		\midrule
		\multirow{4}*{MNIST}&0.1\%   &66\% &84\%
		\\
		~& 0.2\%  &68\% &88\%  \\
		~& 0.5\%  &72\%  &92\%\\
		~ & 1\%  &\textbf{75\%}&\textbf{99\%}\\
		\midrule
		\multirow{4}*{CIFAR-10}	&0.1\%    &58\%&75\%\\
		~& 0.2\%  &59\%&78\%\\
		~& 0.5\%  &63\%&84\%\\
		~ & 1\%   &\textbf{70\%}&\textbf{94\%}\\
		\bottomrule
	\end{tabular}
\end{table}

\begin{table}[!t]
	\caption{Poisoning performance of other poisoning attacks.}
	\label{tab-6}
	\centering
	\begin{tabular}{ccccc}
		\toprule
		Dataset & Poisoning Attack  &$\Delta$\\
		\midrule
		\multirow{4}*{MNIST}&Pseudo Label Flipping Poisoning   &5.76\%
		\\
		~& Gaussian Attack  &12.34\%\\
		~& Adversarial Data Poisoning  &3.69\%\\
		\midrule
		\multirow{4}*{CIFAR-10}  & Pseudo Label Flipping Poisoning  &6.47\%
		\\
		~& Gaussian Attack  &12.69\%\\
		~& Adversarial Data Poisoning  &4.98\%\\
		\bottomrule
	\end{tabular}
\end{table}

\begin{table}[!t]
	\caption{Poisoning Performance with different SSL Methods on MNIST dataset.}
	\label{tab-3}
		\centering
	\begin{tabular}{ccccc}
		\toprule
		Learning Method& $p$  & ASR (MNIST) &ASR (CIFAR-10)\\
		\midrule
		\multirow{4}*{FixMatch}&0.1\%   &66\% &58\%
		\\
		~& 0.2\% &68\%&59\%  \\
		~& 0.5\%  &72\% &63\%\\
		~ & 1\%  &\textbf{75\%}&\textbf{70\%}\\
		\midrule
		\multirow{4}*{MixMatch}	&0.1\%  &62\%&57\%\\
		~& 0.2\%  &65\%&62\%\\
		~& 0.5\%  &68\%&63\%\\
		~ & 1\%  &\textbf{71\%}&\textbf{68\%}\\
		\bottomrule
	\end{tabular}
\end{table}

\begin{table}[!t]
	\caption{Poisoning Performance under Non-i.i.d. Setting.}
	\label{tab-5}
		\centering
	\begin{tabular}{cccccc}
		\toprule
		Dataset & $p$&Learning Method & datasetting  & ASR\\
		\midrule
		\multirow{4}*{MNIST}&0.1\%&\multirow{4}*{FixMatch}  & \multirow{4}*{non-i.i.d.} &96\%
		\\
		~&0.2\%& ~ & ~  &97\%\\
		~& 0.5\%&~ &~  &97\%\\
		~ & 1\%&~ & ~  &\textbf{100\%}\\
		\midrule
		\multirow{4}*{CIFAR-10}&0.1\%&\multirow{4}*{FixMatch}  & \multirow{4}*{non-i.i.d.} &86\%
		\\
		~& 0.2\%&~ & ~ &88\%\\
		~& 0.5\%&~ &  ~  &94\%\\
		~ & 1\%&~  &  ~ &\textbf{99\%}\\
		\bottomrule
	\end{tabular}
\end{table}

\subsection{Defense Performance}
\begin{figure}[!t]
	\centering
	\large
	\subfigure [MNIST i.i.d. dataset]{\includegraphics[width=0.45\linewidth]{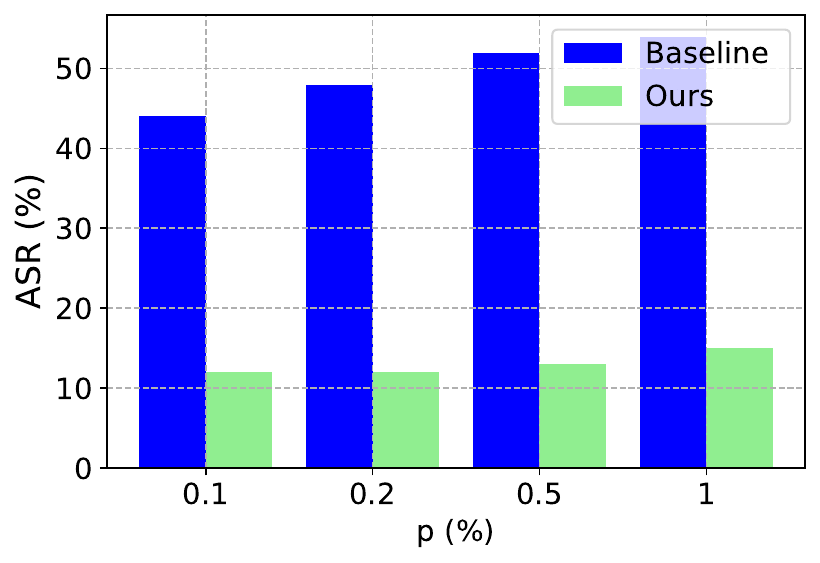}
		\label{fig-a}}
	\hfill
	\subfigure[CIFAR-10 i.i.d. dataset]{	\includegraphics[width=0.45\linewidth]{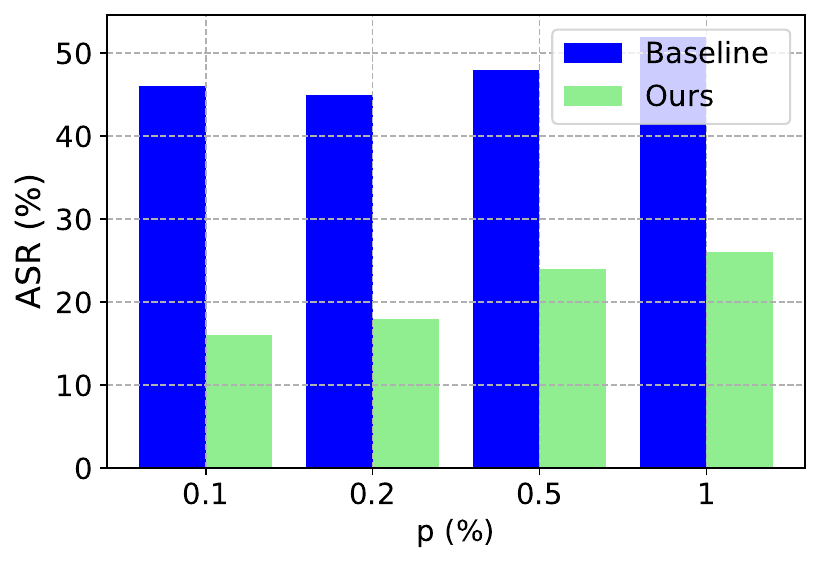}
		\label{fig-b}}
	\caption{ASR of the proposed defense and the baseline defense under a consistency loss poisoning attack using using FixMatch.}
	\label{fig-10}
\end{figure}

\begin{figure}[!t]
	\centering
	\large
	\subfigure [MNIST i.i.d. dataset]{\includegraphics[width=0.45\linewidth]{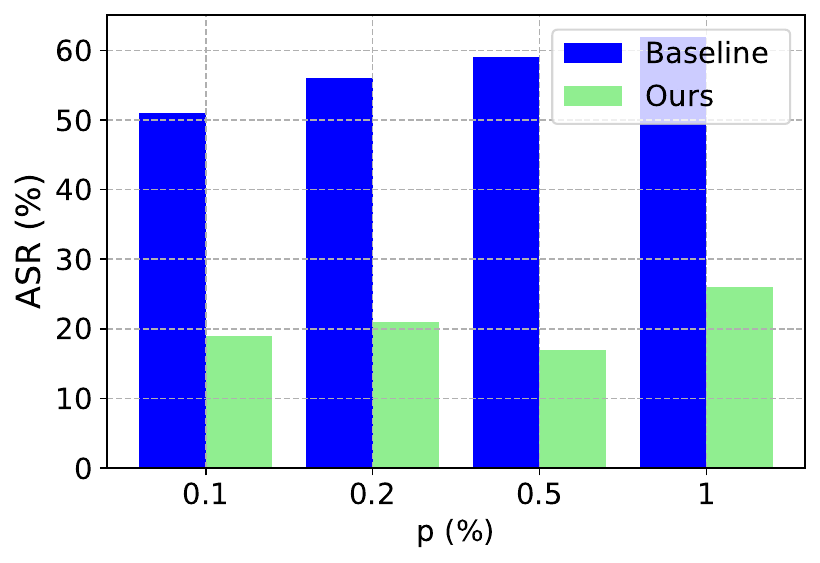}
		\label{fig-a-1}}
	\hfill
	\subfigure[CIFAR-10 i.i.d. dataset]{	\includegraphics[width=0.45\linewidth]{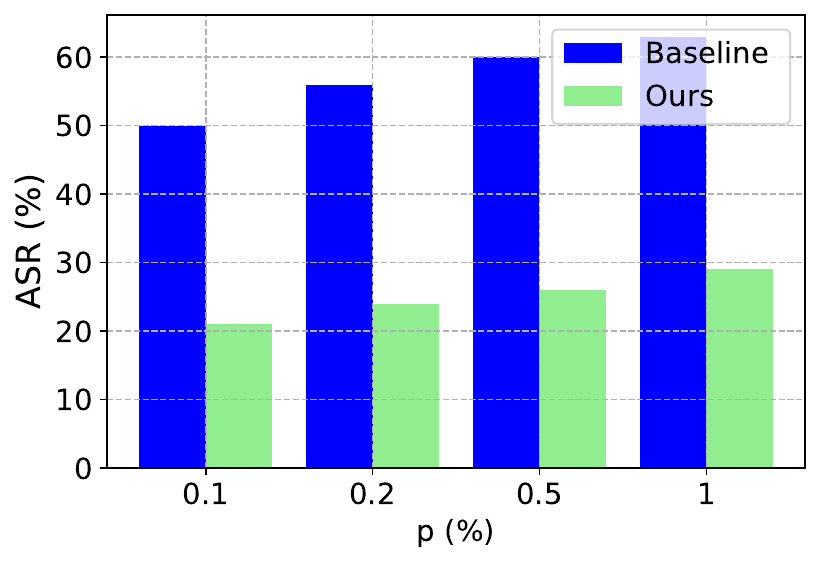}
		\label{fig-b-1}}
	\caption{ASR of the proposed defense and the baseline defense under a consistency loss poisoning attack using MixMatch.}
	\label{fig-11}
\end{figure}

\noindent \textbf{Defend against Our Proposed Consistency Loss Poisoning Attacks.} In this section, we comprehensively evaluate the proposed defense. In this experiment, we focus on distributed poisoning scenarios. We fixed the poisoning rate $p=1\%$ and conducted experiments under the i.i.d. datasetting. Note that other parameter settings and semi-supervised learning method settings are the same as those in Sec. \ref{sec-5-1}. First, the proposed client selection strategy relies on the optimal threshold hyperparameter $\alpha$ to identify malicious clients. Thus, we explore the impact of different threshold $\alpha$ on the performance of the proposed system. For the MNIST dataset, the system performance is the best when $\alpha=0.90$; for the CIFAR-10 dataset, the system performance is the best when $\alpha=0.85$. The reason is that the model updates generated on different datasets (using different CNN models) are different, so one needs to set different thresholds for different datasets. First, for our proposed consistency loss poisoning attack, we compare the performance of the proposed defense with a baseline scheme. Here, we use SSFL with Byzantine-robust aggregation rule (i.e., Geometric Median Aggregation ($\mathrm{GMA}$ rule), please refer to Appendix \ref{app-3} and Reference \cite{ref-72}) as the baseline scheme for comparison of our defense. $\mathrm{GMA}$ is a popular and commonly used Byzantine robust aggregation method and has excellent performance against poisoning attacks. In this experiment, we explore the performance of our defense under different $p$ and different semi-supervised methods on two datasets. As shown in Fig. \ref{fig-10} and Fig. \ref{fig-11}, our defense can better mitigate consistency loss poisoning attack under different settings. Specifically, for the CIFAR-10 i.i.d. dataset, when $p=0.1\%$ and the FixMatch method is used, the ASR is reduced to 12\%. Our defense enables the model to obtain high-quality data-label pairs, which makes it difficult for the adversary to design a poisoning path to poison unlabeled data.

\begin{figure*}[!t]
	\centering
	\large
	\subfigure [MNIST i.i.d. dataset]{\includegraphics[width=0.23\linewidth]{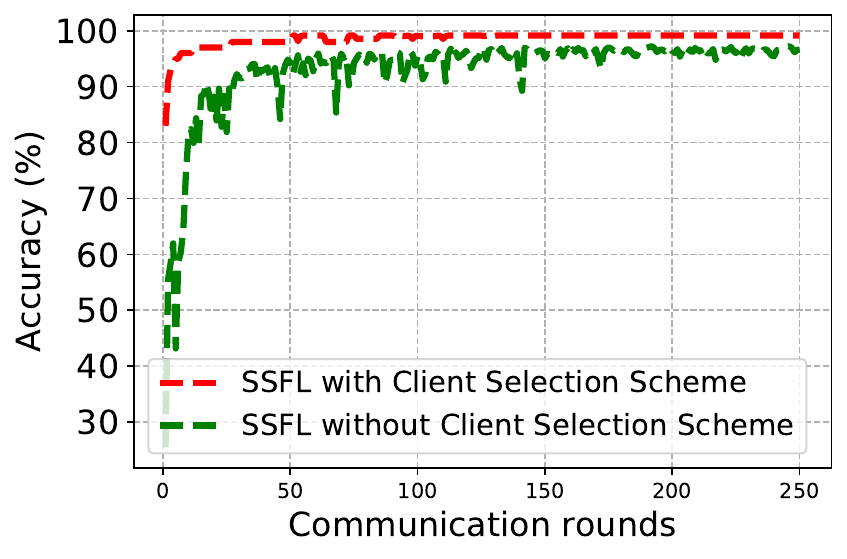}
		\label{fig-6-a}}
	\hfill
	\subfigure[MNIST non-i.i.d. dataset]{	\includegraphics[width=0.23\linewidth]{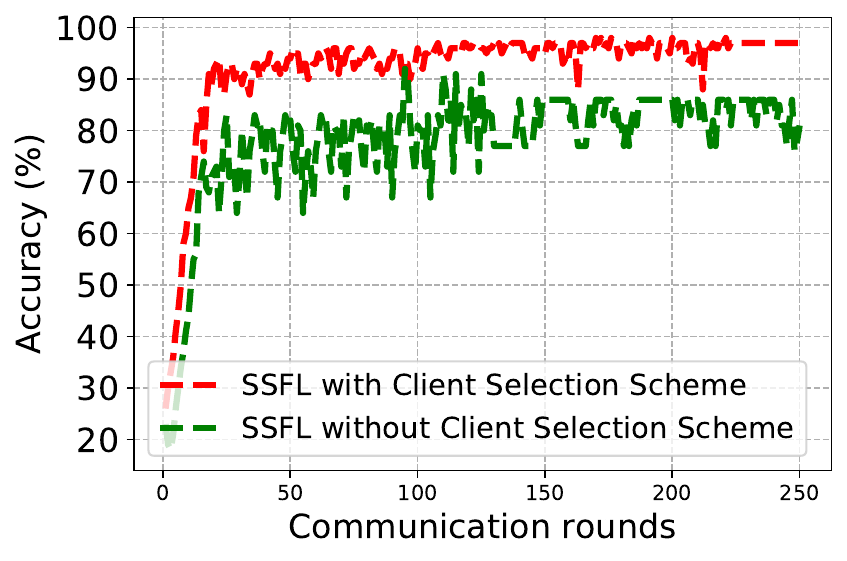}
		\label{fig-6-b}}
	\hfill
	\subfigure [CIFAR-10 i.i.d. dataset]{\includegraphics[width=0.23\linewidth]{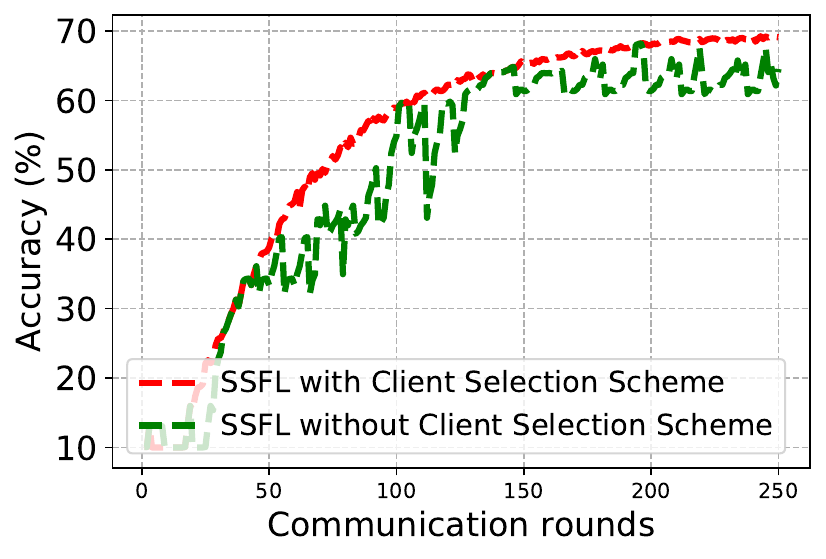}
		\label{fig-6-c}}
	\hfill
	\subfigure[CIFAR-10 non-i.i.d. dataset]{	\includegraphics[width=0.23\linewidth]{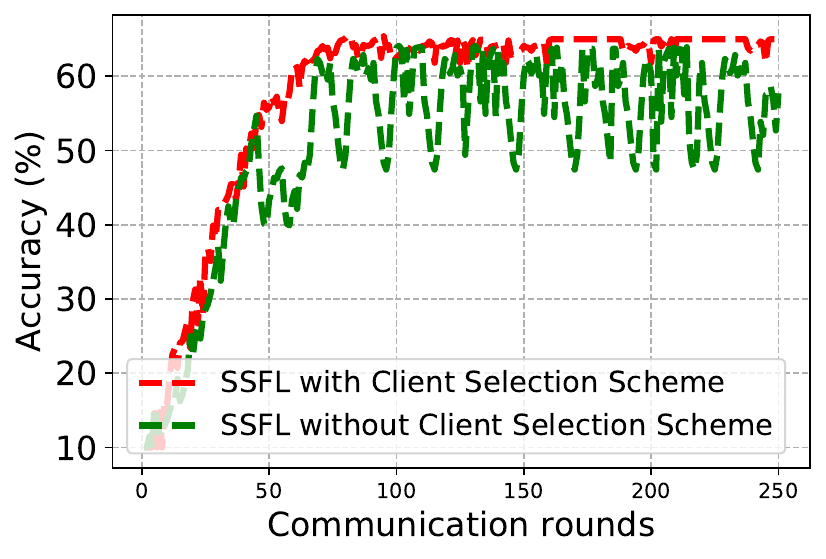}
		\label{fig-6-d}}
	\caption{Performance of the SSFL with and without the proposed defense on pseudo label-flipping attacks.}
	\label{fig-6}
	\vspace{-0.5cm}
\end{figure*}

\begin{figure*}[!t]
	\centering
	\large
	\subfigure [MNIST i.i.d. dataset]{\includegraphics[width=0.23\linewidth]{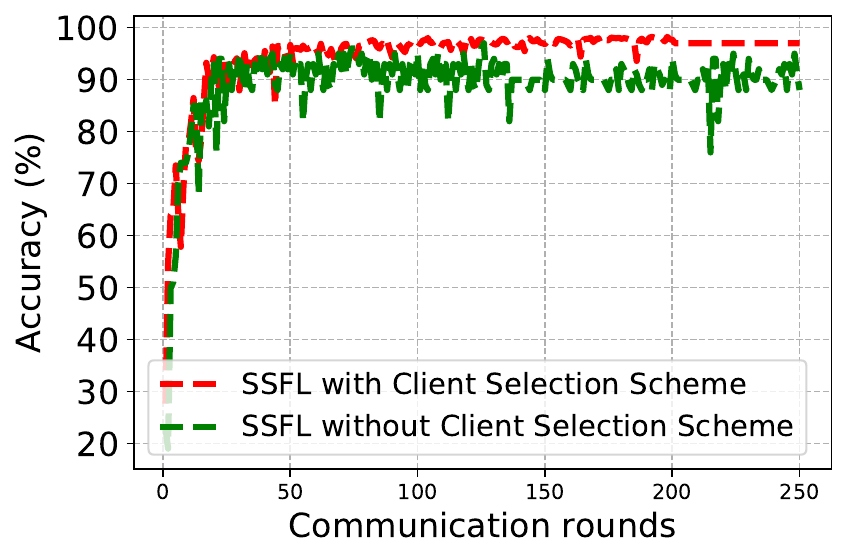}
		\label{fig-7-a}}
	\hfill
	\subfigure[MNIST non-i.i.d. dataset]{	\includegraphics[width=0.23\linewidth]{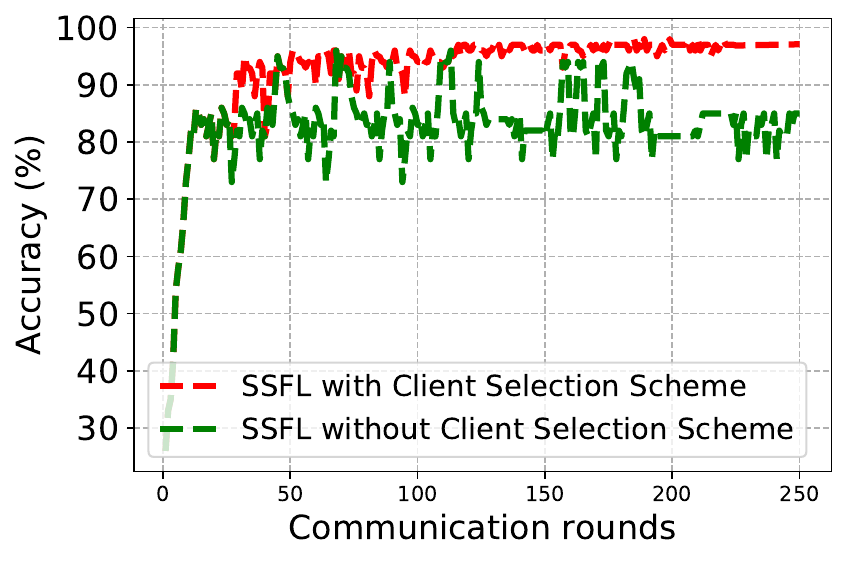}
		\label{fig-7-b}}
	\hfill
	\subfigure [CIFAR-10 i.i.d. dataset]{\includegraphics[width=0.23\linewidth]{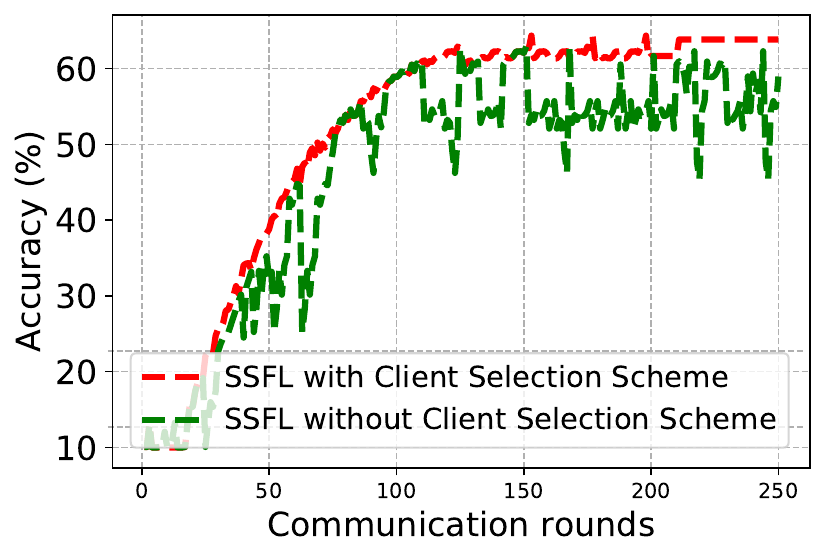}
		\label{fig-7-c}}
	\hfill
	\subfigure[CIFAR-10 non-i.i.d. dataset]{	\includegraphics[width=0.23\linewidth]{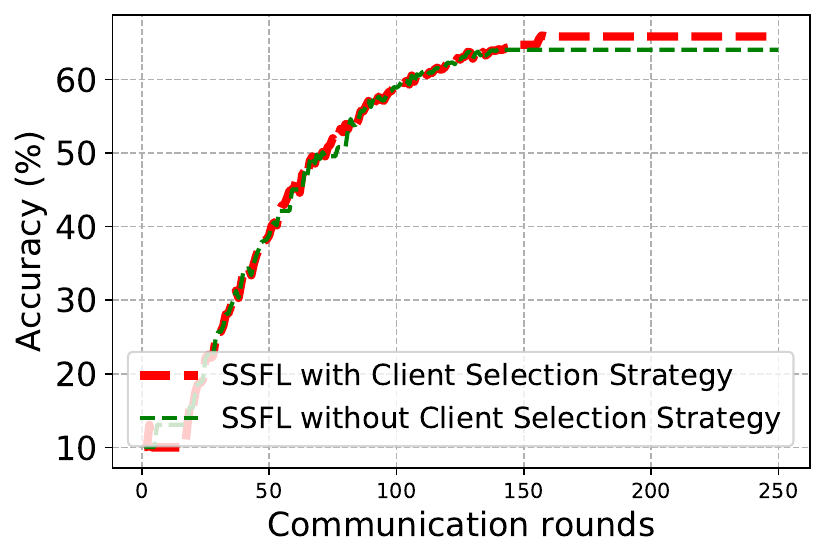}
		\label{fig-7-d}}
	\caption{Performance of the SSFL with and without the proposed defense on Gaussian attacks.}
	\label{fig-7}
		\vspace{-0.5cm}
\end{figure*}

\begin{figure*}[!t]
	\centering
	\large
	\subfigure [MNIST i.i.d. dataset]{\includegraphics[width=0.23\linewidth]{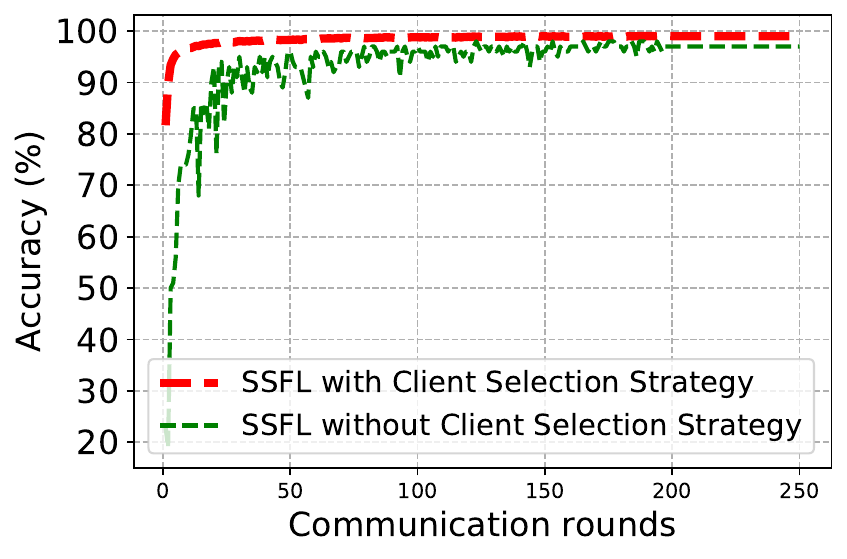}
		\label{fig-8-a}}
	\hfill
	\subfigure[MNIST non-i.i.d. dataset]{	\includegraphics[width=0.23\linewidth]{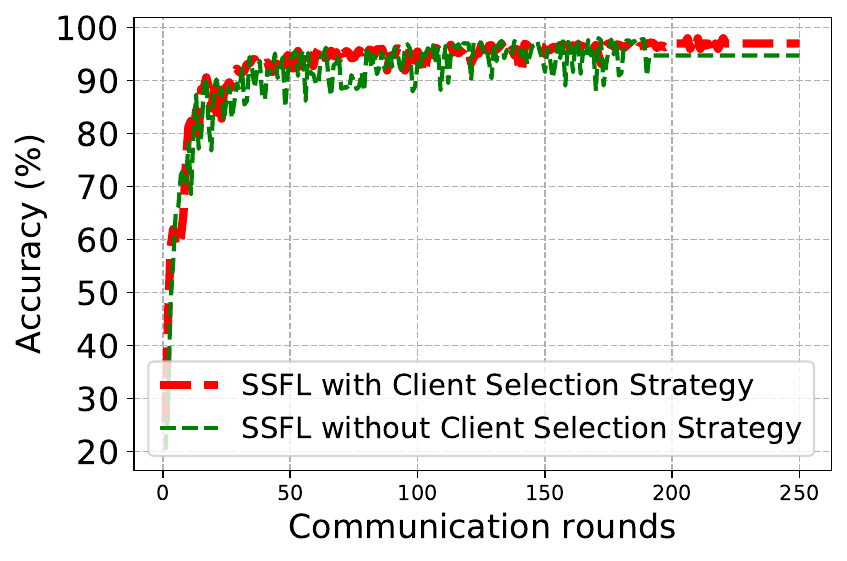}
		\label{fig-8-b}}
	\hfill
	\subfigure[CIFAR-10 i.i.d. dataset]{\includegraphics[width=0.23\linewidth]{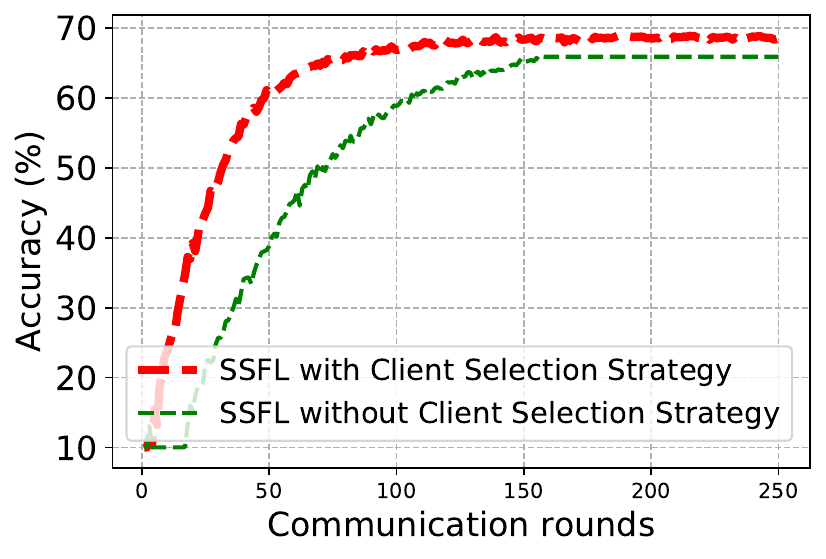}
		\label{fig-8-c}}
	\hfill
	\subfigure[CIFAR-10 non-i.i.d. dataset]{	\includegraphics[width=0.23\linewidth]{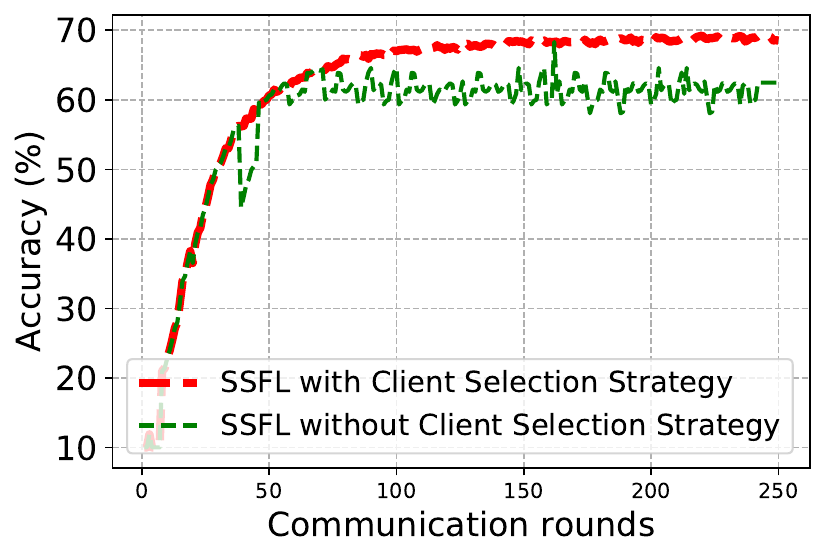}
		\label{fig-8-d}}
	\caption{Performance of the SSFL with and without the proposed defense on adversarial data poisoning attacks.}
	\label{fig-8}
		\vspace{-0.5cm}
\end{figure*}

\noindent \textbf{Defend against Other Poisoning Attacks.} Second, we evaluate the performance of the proposed defense against three typical poisoning attacks (i.e., pseudo label-flipping attacks, Gaussian attacks, and adversarial data poisoning attacks) on the MNIST and CIFAR-10 datasets. Here, we assume the number of compromised clients $c=30$, which is consistent with prior art \cite{ref-13}. These compromised clients can collude with each other to launch attacks. Meanwhile, our experiment randomly selects $\frac{1}{3}$ of all malicious clients in each round to demonstrate the sophisticated attack strategy in practice. In this context, we evaluate the performance of the proposed defense against pseudo label-flipping attacks and Gaussian attacks on the MNIST and CIFAR-10 datasets. In Fig. \ref{fig-6-a}--Fig. \ref{fig-6-d} and Fig. \ref{fig-7-a}--Fig. \ref{fig-7-d}, we show the performance comparison between the SSFL with the proposed client selection strategy and the SSFL without it (i.e., using $\mathrm{GMA}$ scheme) on the i.i.d. and non-i.i.d. MNIST dataset and CIFAR-10 dataset. Specifically, both methods can effectively defend against pseudo label-flipping and Gaussian attacks. From the experimental results, the performance of the proposed defense is better than the baseline scheme. The reason is that in our defense, the server selects the benign clients by comparing the cosine similarity and distribution similarity between the clients' updates and its own updates, which prevents malicious clients from participating in SSFL training. 

Then we evaluate the performance of the proposed defense against adversarial data poisoning attacks on the MNIST and CIFAR-10 datasets. In this attack, compromised clients collude with each other to launch a targeted local model poisoning attack. Specifically, the adversary can access the local dataset to generate the adversarial dataset $\mathcal{D}_{adv}$ and then make all compromised clients collude with each other to maximize  Eq. \eqref{eq-11}. We show the accuracy evaluation results on the i.i.d.. and non-i.i.d.. of the MNIST dataset in Fig. \ref{fig-8-a} and Fig. \ref{fig-8-b}, respectively, which demonstrate that the designed client selection strategy can help the SSFL maintain accuracy and convergence in the presence of adversarial attacks. We can find the same results in Fig. \ref{fig-8-c} and Fig. \ref{fig-8-d} for the CIFAR-10 dataset. It can be seen that the baseline using Byzantine-robust aggregation rules cannot resist the adversarial attacks, and especially it cannot achieve convergence under the non-i.i.d.. setting. The reason is that the system we designed uses minimizing $\ell_1$ norm and $\ell_2$ norm sparsifies local model updates by to alleviate the adverse effects of non-i.i.d.. settings on the global model. 

Furthermore, the proposed $\mathrm{QRA}$ aggregation rule can increase the contribution of clients holding high-quality local updates to improve performance. On the other hand, as shown in the experimental results, adversarial attacks have a stronger ability to poison the global model than label-flipping attacks and Gaussian attacks. This is because adversarial poisoning attacks are concealed and do not affect the convergence of the global model. Therefore, choosing a client with high-quality model updates is necessary to defend against such attacks.

\section{Conclusion}\label{sec-6}
Semi-supervised federated learning no longer relies on a large amount of labeled data to make FL practical. However, it provides a new attack surface for the adversary. This paper revealed that SSFL is vulnerable to poisoning attacks in practice, especially by poisoning unlabeled data. Specifically, we designed a powerful and covert poisoning attack that allows adversaries to poison the SSFL by injecting unlabeled data. Notably, the insight of our attack uses the inherent advantages of semi-supervised training loss to enable the model output target label on poisoned unlabeled data in a self-fooling manner. Experimental results show that the adversary just needs to manipulate 0.1\% of the training data to achieve powerful attack performance. This implies that we need to tailor a defensive strategy for the SSFL in practice. To this end, our defense first departs from prior works by designing a minimax optimization-based client selection strategy and the quality-based aggregation rule, which enjoys the advantages of high-quality label information and updates aggregation as well as the ability to remove the poisoned label without compromising performance. We conducted an extensive evaluation over popular benchmark datasets, and the results validated the practical performance of our defense. Besides, our defense is robust to poisoning attacks under i.i.d.. and non-i.i.d.. settings.

\bibliographystyle{IEEEtran}
 
\bibliography{sample-base}

\appendix

\section{Geometric Median Aggregation}\label{app-3}
\begin{figure}[!t]
	\centering
	\includegraphics[width=0.9\linewidth]{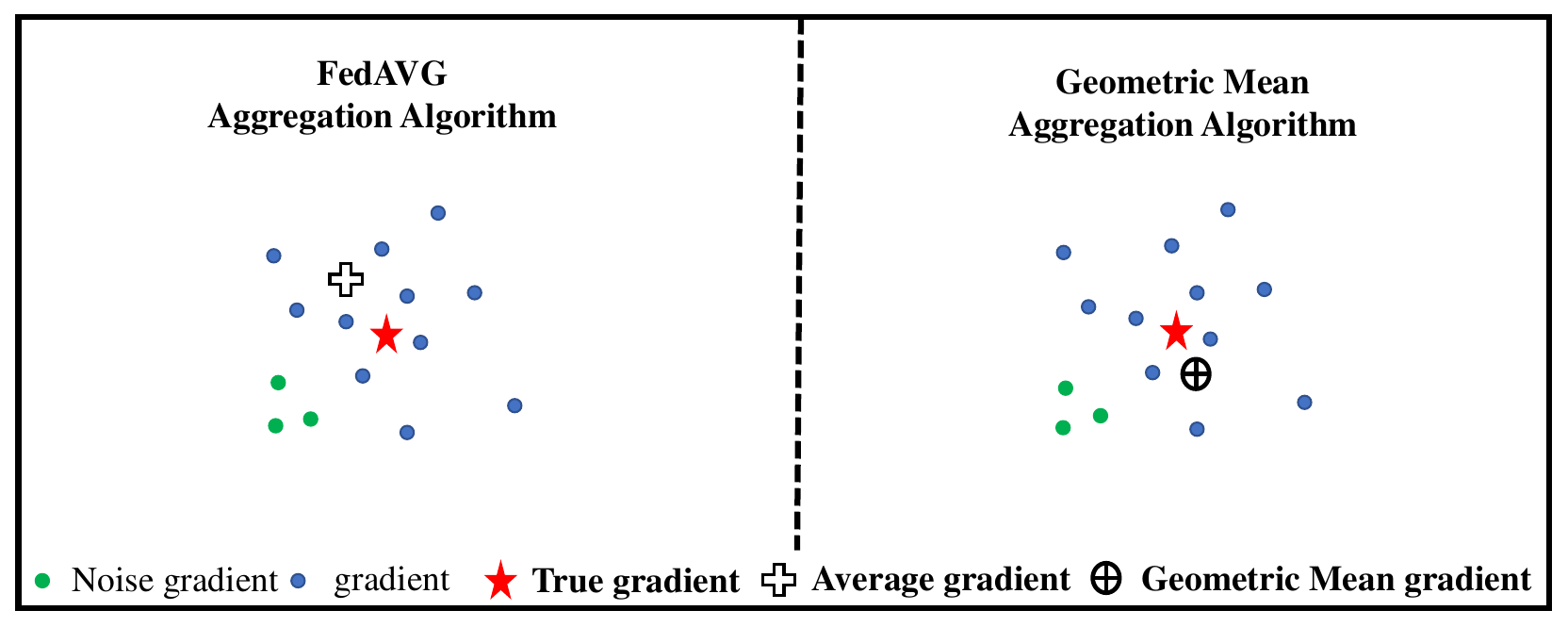}
	\caption{An overview of the comparison between FedAvg and geometric median aggregation algorithms.}
	\label{fig-app}
\end{figure}
In this paper, we use $\mathrm{GMA}$ \cite{ref-72} defense as our baseline. Next, we introduce in detail the core ideas of the $\mathrm{GMA}$ defense method. Thus, we give the definition of $\mathrm{GMA}$ as follows:
\begin{definition}
\textbf{Geometric Median Aggregation ($\mathrm{GMA}$):} Let $\{ z,z \in Z\} $ be a subset of the space of natural numbers, thus, the geometric median is defined as follows:
\begin{equation}
	\mathop {{\mathrm{geomed}}}\limits_{z \in Z} \{ z\} : = \arg \mathop {\min }\limits_y \sum\limits_{z \in Z} {{{||}}y - z||}.
\end{equation}
Referring to FedAvg \cite{ref-5}, the formal definition of geometric median aggregation is as follows:
\begin{equation}\label{eq-11}
{\omega _{k + 1}} = {\omega _k} + \frac{1}{m}\sum\limits_{i = 1}^m {\mathop {{\mathrm{geomed}}}\limits_{\omega  \in {\mathbb{R}^d}} \{ {\omega _i}\} } .
\end{equation}
\end{definition}

Here, we address the following key questions:

\noindent 1) Why $\mathrm{GMA}$ can resist poisoning attacks?

\noindent 2) Why $\mathrm{GMA}$ is better than FedAvg aggregation algorithm?

To answer the above queations, we assume that there are some noisy gradients. As shown in Fig. \ref{fig-app}, if there are some noisy gradients, the average gradient produced by the FedAvg aggregation algorithm is quite different from the true gradient. On the contrary, the geometric median aggregation algorithm is robust to the noisy gradients and the aggregation result is closer to the true gradient. Second, FedAvg aggregation algorithm is susceptible to noisy updates. The main reason is that the stochastic gradient descent optimizer is very sensitive to random noise in the process of gradient descent\cite{ref-37}. Noisy updates cause the server to get incorrect aggregation results and harm the performance of the system.

\end{document}